\documentclass[journal]{IEEEtran}
\usepackage{mdframed}
\usepackage{amsmath}
\usepackage{amssymb}
\usepackage{amsthm}
\usepackage{graphicx}
\usepackage{enumerate}
\usepackage{blindtext}
\usepackage{multirow}
\usepackage{multicol}
\usepackage{placeins}
\usepackage{hyperref}
\usepackage{caption}
\usepackage{epstopdf}
\usepackage{algorithm}
\usepackage{algpseudocode}
\newcommand*\rot[1]{\rotatebox{90}{#1}}
\usepackage{mathtools}
\usepackage[switch,columnwise]{lineno} 
\newcommand{\defeq}{\vcentcolon=}
\DeclareMathOperator\erfc{erfc}

\DeclareMathOperator\vectx{\vect{x}}
\DeclareMathOperator\vx{\vect{x}}

\DeclareMathOperator\vw{\vect{w}}
\DeclareMathOperator\vecty{\vect{y}}
\DeclareMathOperator\vectg{\vect{\gamma}}

\DeclareMathOperator*{\argmin}{arg\,min}
\DeclareMathOperator*{\argmax}{arg\,max}
\newcommand{\vect}[1]{\boldsymbol{#1}}
\DeclareMathOperator*{\inv}{^{-1}}

\title{Rectified Gaussian Scale Mixtures and the Sparse Non-Negative Least Squares Problem}
\author{Alican~Nalci,~\IEEEmembership{Student Member,~IEEE,}
		    Igor~Fedorov,~\IEEEmembership{Student Member,~IEEE,}
		    Maher~Al-Shoukairi,~\IEEEmembership{Student Member,~IEEE,}
		    Thomas~T.~Liu,~\IEEEmembership{Member,~IEEE,}
		    and Bhaskar~D.~Rao~\IEEEmembership{Fellow,~IEEE}

\thanks{Alican Nalci, Igor Fedorov, Maher Al-Shoukairi and Bhaskar D. Rao  are with the Department of Electrical and Computer Engineering, University of California, San Diego, 9500 Gilman Drive, La Jolla, CA 92093, USA. (Correspondence: analci@ucsd.edu)

Thomas T. Liu is with the Departments of Radiology,  Psychiatry and Bioengineering, and UCSD Center for Functional MRI, University of California, San Diego, 9500 Gilman Drive, La Jolla, CA 92093, USA

We would like to thank Mr. Sung-En Chiu for his comments on an earlier version of this manuscript. 

This work was partially supported by NIH grant R21MH112155.}}
\begin{document}
\maketitle
\begin{abstract}
In this paper, we develop a Bayesian evidence maximization framework to solve the sparse non-negative least squares (S-NNLS) problem. 
We introduce a family of probability densities referred to as the Rectified Gaussian Scale Mixture (R-GSM) to model the sparsity enforcing prior distribution for the solution.
The R-GSM prior encompasses a variety of heavy-tailed densities such as the rectified Laplacian and rectified Student-t distributions with a proper choice of the mixing density. 
We utilize the hierarchical representation induced by the R-GSM prior and develop an evidence maximization framework based on the Expectation-Maximization (EM) algorithm.
Using the EM based method, we estimate the hyper-parameters and obtain a point estimate for the solution.
We refer to the proposed method as rectified sparse Bayesian learning (R-SBL).
We provide four R-SBL variants that offer a range of options for computational complexity and the quality of the E-step computation.
These methods include the Markov chain Monte Carlo EM, linear minimum mean-square-error estimation, approximate message passing and a diagonal approximation. 
Using numerical experiments, we show that the proposed R-SBL method outperforms existing S-NNLS solvers in terms of both signal and support recovery performance, and is also very robust against the structure of the design matrix.
\end{abstract}
\begin{IEEEkeywords}
Non-negative least squares, Sparse Bayesian learning, Sparse signal recovery, rectified Gaussian scale mixtures
\end{IEEEkeywords}

\section{Introduction}
\label{sec:intro}
\IEEEPARstart{T}{his} work considers the following signal model
\begin{align}\label{intro_1}
\vecty= \vect\Phi\vectx +\vect{v},
\end{align}
where the solution vector $\vect{x} \in {\mathbb{R}_+^{M}}$ is assumed to be non-negative, the matrix $\vect{\Phi} \in \mathbb{R}^{N\times M}$ is fixed and obtained from the physics of the underlying problem, $\vect{y} \in \mathbb{R}^{N}$ is the measurement, and $\vect{v}$ is the additive noise modeled as a zero mean Gaussian with uncorrelated entries $v_i \sim \mathcal{N}(0,\sigma^2)$. 

Recovering $\vect{x}$ using the signal model in Eq. \eqref{intro_1} is known as solving the non-negative least squares (NNLS) problem. 
NNLS has a rich history in the context of methods for solving systems of linear equations \cite{lawson1974solving}, density estimation \cite{jedynak2005maximum}, and non-negative matrix factorization (NMF) \cite{peharz2012sparse,kim2007sparse,kim2008nonnegative,fedorov2018unified}. NNLS is also widely used in text mining \cite{pauca2004text}, image hashing \cite{monga2007robust}, speech enhancement \cite{loizou2005speech}, spectral decomposition \cite{fevotte2009nonnegative}, magnetic resonance chemical shift imaging \cite{sajda2004nonnegative}, and impulse response estimation \cite{lin2006bayesian}.

The maximum-likelihood solution for the signal model in Eq. \eqref{intro_1} is given by
\begin{equation}
\begin{aligned}\label{intro_2}
& \underset{\vectx \geq \vect0}{\text{minimize}}
& \Vert\vecty -\vect\Phi\vectx\Vert_2.
\end{aligned}
\end{equation}

In many applications, $N<M$ and Eq. \eqref{intro_1} is under-determined.
This means that a unique solution for $\vect{x}$ may not exist. 
Recovering a unique solution is possible if more information is known \textit{a-priori} about the solution vector. 
For example, a useful assumption is that the solution vector is \textit{sparse} and contains only a few non-zero elements \cite{potter2010sparsity,hurmalainen2012group,lustig2006kt}.
In this case, the sparsest solution (assuming a noiseless case) can be recovered by modifying Eq. \eqref{intro_2} to
\begin{equation}
\begin{aligned}\label{intro_3}
& \underset{\vectx \geq \vect0, ~\vecty  = \Phi\vectx}{\text{minimize}}
& \Vert\vectx\Vert_0,
\end{aligned}
\end{equation}
where $\Vert . \Vert_0$ is the $\ell_0$ pseudo-norm, which counts the non-zero elements in $\vectx$. 
The count of non-zero elements is also referred to as the cardinality of the solution.
Then, the recovery objective in Eq. \eqref{intro_3} is to minimize the cardinality of $\vectx$ while satisfying the optimization constraints. 
This approach is commonly referred to as solving the sparse NNLS (S-NNLS) problem.

The S-NNLS problem is becoming increasingly popular in certain applications where the non-negative solution needs to be recovered from a limited number of measurements.
For example, in \cite{ghosh2013fiber} an S-NNLS method was applied to magnetic resonance imaging (MRI) data to reconstruct narrow fiber-crossings from a limited number of acquisitions. 
In \cite{meng2011bayesian}, another method was used to uncover regulatory networks from micro-array mRNA expression profiles from breast cancer data. 
In \cite{nalci:2016ISMRM,liu2017global}, an S-NNLS method was applied to functional MRI data to estimate sparsely repeating spatio-temporal activation patterns in the human brain.
S-NNLS solvers are also used in applied mathematics for designing dictionaries for sparse representations, such as sparse NMF and non-negative K-SVD \cite{peharz2012sparse,aharon2005k}.

The objective function in Eq. \eqref{intro_3} is not tractable since the $\ell_0$ penalty is not convex and the problem is NP-hard \cite{jiang2009note,eladbook}. 
Therefore, `greedy' algorithms have been proposed to approximate the solution \cite{tropp2007signal,needell2009cosamp,mallat1993matching,needell2009uniform,c3}. 
An example is the class of algorithms known as Orthogonal Matching Pursuit (OMP) \cite{ tropp2007signal,pati1993orthogonal}, which greedily selects the non-zero elements of $\vect{x}$.
In order to adapt OMP to the S-NNLS problem, the criterion by which a new non-zero element of $\vectx$ is selected is modified to select the one having the largest \textit{positive} value \cite{c3}. 

Another approach in this class of algorithms finds an $\vectx$ such that $\Vert \vecty - \Phi\vectx \Vert_2 \leq \epsilon$ and $\vectx \geq 0$ using the active-set Lawson-Hanson algorithm \cite{lawson1974solving} and then prunes $\vectx$ until $\Vert \vectx \Vert_0 \leq K$, where $K$ is a pre-specified cardinality \cite{peharz2012sparse}.

Greedy algorithms are computationally attractive but may lead to sub-optimal solutions. 
Therefore, convex relaxations of the $\ell_0$ penalty have been 
proposed \cite{eladbook,efron2004least,figueiredo2007gradient,wright2009sparse,donoho2005sparse}. 
One simple alternative replaces the $\ell_0$ norm with the $\ell_1$ norm and reformulates the problem in Eq. \eqref{intro_3} as
\begin{equation}
\begin{aligned}\label{nnl1}
& \underset{\vectx \geq \vect0}{\text{minimize}}
& & \Vert \vecty - \vect{\Phi}\vectx\Vert_2 + \lambda \Vert \vectx \Vert_1,
\end{aligned}
\end{equation}
where $\lambda >  0$ is a regularization parameter to account for the measurement noise. 
The advantage of the formulation in Eq. \eqref{nnl1} is that it is a convex optimization problem and can be solved by a number of methods \cite{donoho2005sparse,nocedal2006numerical,boyd2004convex,khajehnejad2011sparse}. 
One approach is to estimate $\vectx$ with projected gradient descent \cite{lin2007projected}.

In fact, the $\ell_1$ penalty in Eq. \eqref{nnl1} can be replaced by any arbitrary sparsity inducing surrogate function $g(\vectx)$, thus leading to alternative methods based on solving
\begin{equation}
\begin{aligned}\label{nnls g}
& \underset{\vectx \geq \vect0}{\text{minimize}}
& & \Vert \vecty - \vect{\Phi}\vectx\Vert_2 + \lambda g(\vectx).
\end{aligned}
\end{equation}
For example, a surrogate $g(\vectx) = \sum_{i=1}^M \log \left( x_i^2 + \beta \right)$ leads to an iterative reweighted optimization approach \cite{grady2008compressive,chartrand2008iteratively}.

A promising view on the S-NNLS problem is to cast the entire problem in a Bayesian framework and consider the maximum a-posteriori (MAP) estimate of $\vectx$ given $\vecty$
\begin{align}\label{bayes}
\vect{x}_{MAP} &= \argmax_{\vect{x}} p(\vect{x} | \vect{y}).
\end{align}

There is a strong connection between the MAP framework and the previous deterministic formulations. 
Recently, it has been shown that formulations of the form in Eq. \eqref{nnls g} can be represented by using the formulation in Eq. \eqref{bayes} with a proper choice of $p(\vectx)$ \cite{giri2015type}. 
For example, considering a separable $p(\vectx)$ of the form
\begin{align}\label{seperable prior}
p(\vect{x}) = \prod_{i=1}^M p(x_i),
\end{align}
the $\ell_1$ regularization approach in Eq. \eqref{nnl1} (i.e. a choice of $g(\vectx)=\Vert \vectx \Vert_1$ in Eq. \eqref{nnls g}) is equivalent to the Bayesian formulation in Eq. \eqref{bayes} with an
exponential prior for $x_i$. In this work our emphasis will be on Bayesian approaches for solving Eq. \eqref{intro_1}. 
\subsection{Contributions of the paper}
\begin{itemize}
\item We introduce a family of non-negative probability densities referred to as the rectified Gaussian scale mixture (R-GSM) to model non-negative and sparse solutions.
\item We discuss how the R-GSM prior encompasses other sparsity inducing non-negative priors, such as the rectified Laplacian and rectified Student-t distributions through a proper choice of the mixing density.
\item We detail how the R-GSM prior can be utilized to solve the S-NNLS problem using an evidence maximization based estimation procedure that utilizes the expectation-maximization (EM) framework. 
We refer to this technique as rectified sparse Bayesian learning (R-SBL).
\item We provide four alternative R-SBL methods that offer a range of options for computational complexity and the quality of the E-step computation. These methods include the Markov Chain Monte Carlo EM, linear minimum mean-square-error estimation, approximate message passing and a diagonal approximation. 
\item We use extensive empirical results to show the robustness and superiority of the R-GSM priors and R-SBL algorithm for the S-NNLS problem. 
Especially, under various i.i.d. and non-i.i.d. settings for the design matrix $\mathbf{\Phi}$.
\end{itemize}

\subsection{Organization of the paper}
In Section \ref{sec:rgsm}, we discuss the advantages of using scale mixture priors for $p(\vectx)$ and introduce the R-GSM prior.
In Section \ref{sec:inference}, we define the Type I and Type II Bayesian approaches to solve the S-NNLS problem and introduce the R-SBL framework.
We provide the details of an evidence maximization based estimation procedure in Section \ref{sec:type2}.
We present empirical results comparing the proposed R-SBL algorithm to the baseline S-NNLS solvers in Section \ref{sec:results}. 

\section{Rectified Gaussian Scale Mixtures}
\label{sec:rgsm}
We assume separable priors of the form in Eq. \eqref{seperable prior} and focus on the choice of $p(x_i)$.
The choice of prior plays a central role in the Bayesian inference \cite{tipping2001sparse,babacan2010bayesian,ji2008bayesian}.
For the S-NNLS problem, the prior must induce sparsity and satisfy the non-negativity constraints.
Consequently, we consider the hierarchical scale mixture prior
\begin{align}
\label{eq:scale mixture}
p(x_i) = \int_0^\infty p(x_i | \gamma_i) p(\gamma_i) d\gamma_i.
\end{align}

The scale mixture prior was first considered in the form of Gaussian Scale Mixtures (GSM) with $p(x_i | \gamma_i) = \mathcal{N}(x_i; 0, \gamma_i)$ \cite{andrews1974scale}. 
Super-gaussian densities are suitable priors for promoting sparsity \cite{tipping2001sparse,wipf2007empirical} and can be represented in the form shown in Eq. \eqref{eq:scale mixture} with a proper choice of mixing density $p(\gamma_i)$  \cite{palmer2006variational,palmer2005variational,lange1993normal,dempster1980iteratively,dempster1977maximum}. This has made scale mixture priors valuable for the standard sparse signal recovery problem. 
Another advantage of the scale mixture prior is that, it establishes a Markovian structure of the form
\begin{align}\label{markovv}
\vectg \rightarrow \vectx \rightarrow \vecty,
\end{align}
where inference can be performed in the $\vectx$ domain (referred to as Type I) and in the $\vectg$ domain (Type II).
Experimental results for the standard sparse signal recovery problem show that performing inference in the $\vectg$ domain consistently achieves superior performance \cite{giri2015type,tipping2001sparse,wipf2011latent,al2018gamp}.

The Type II procedure involves finding a maximum-likelihood (ML) estimate of $\vect{{\gamma}}$ using evidence maximization and approximating the posterior $p(\vectx|\vecty)$ by $p(\vectx|\vecty,\vect{{\gamma}}_{\text{ML}})$.
The performance gains can be understood by noting that $\vectg$ is deeper than $\vectx$ in Eq. \eqref{markovv}, so the influence of errors in performing
inference in the $\vectg$ domain may be diminished \cite{giri2015type,wipf2011latent}. 
Also, $\vectg$ is close enough to $\vecty$ such that meaningful inference about $\vectg$ can still be performed, mitigating the problem of local minima that is more
prevalent when seeking a Type I estimate of $\vectx$ \cite{wipf2011latent}.

Although priors of the form shown in Eq. \eqref{eq:scale mixture} have been used in the compressed sensing literature (where the signal model is identical to Eq. \eqref{intro_1} without the non-negativity constraint) \cite{giri2015type,c6,zhang2011sparse}, such priors have not been extended to solve the S-NNLS problem. 
Considering the findings that the scale mixture prior has been useful for the development of sparse signal recovery algorithms \cite{giri2015type,wipf2011latent,fedorov2017multimodal}, we propose a R-GSM prior for the S-NNLS problem, where $p(x_i | \gamma_i)$ in
Eq. \eqref{eq:scale mixture} is a rectified Gaussian (RG) distribution. 
We refer to the proposed Type II inference framework as R-SBL.

The univariate RG distribution is defined as
\begin{equation}\label{rgg_density}
\begin{split}
\mathcal{N}^{R}(x;\mu,\gamma) = \sqrt{\dfrac{2}{\pi\gamma}}\dfrac{e^{-\dfrac{(x-\mu)^2}{2\gamma}}u(x)}{\erfc\left(-\dfrac{\mu}{\sqrt{2\gamma}}\right)},
\end{split}
\end{equation}
where $\mu$ is the location parameter (and not the mean), $\gamma$ is the scale parameter, 
$u(x)$ is the unit step function, and $\erfc(x)$ is the complementary error function\footnote{$\erfc(x) = \dfrac{2}{\sqrt{\pi}} \int_{x}^{\infty}  e^{-t^2}dt$}.

As noted in previous works \cite{c1,c2}, closed form inference computations using a multivariate RG distribution are tractable only if the location parameter is zero (by effectively getting rid of the $\erfc(.)$ term). 
Although a non-zero $\mu$ could provide a richer class of priors, possibly to model approximately sparse or non-sparse solutions, considering the tractability issues and the potential overfitting problems (twice as many parameters),  we focus on the R-GSM priors with $\mu=0$ to promote \textit{sparse} non-negative solutions. 
It is a pragmatic choice and adequate for the problem at hand. 

When $\mu=0$, the RG density simplifies to
\begin{align}
\mathcal{N}^{R}(x;0,\gamma)= \sqrt{\dfrac{2}{\pi \gamma}}e^{-\dfrac{x^2}{2\gamma}}u(x).
\end{align}
Thus, the R-GSM prior introduced in this work have the form 
\begin{equation}
p(x) = \int_0^\infty \mathcal{N}^{R}(x;0,\gamma) p(\gamma) d\gamma. 
\end{equation}
Different choices of $p(\gamma)$ lead to different options for $p(x)$ and some examples are presented below.
\subsection{R-GSM representation of sparse priors}
\label{sec:example rgsm}
We can utilize the proposed R-GSM framework to obtain a variety of non-negative sparse priors.
For instance, consider the rectified Laplace prior $p(x) = \lambda e^{-\lambda x} u(x)$.
By using an exponential prior for $p(\gamma) = \frac{\lambda^2}{2}e^{-\frac{\lambda^2 \gamma}{2}} u(\gamma)$, we can express $p(x)$ in the R-GSM framework as \cite{figueiredo2003adaptive} 
\begin{align}
p(x) =& 2u(x) \int_0^\infty \mathcal{N}(x | 0,\gamma) \frac{\lambda^2}{2}e^{-\frac{\lambda^2 \gamma}{2}} u(\gamma) d\gamma \\
 =& \lambda e^{-\lambda x} u(x).
\end{align}

Similarly, by considering a $\mbox{Gamma}(a,b)$ distribution for $p(\gamma)$, we obtain a rectified Student-t distribution for $p(x)$ and Eq. \eqref{eq:scale mixture} simplifies to \cite{tipping2001sparse}
\begin{align}
p(x) =& 2 u(x) \int_0^\infty \mathcal{N}(x | 0,\gamma) \frac{\gamma^{a-1}e^{\frac{-\gamma}{b}}}{a^b \mathsf{\Gamma(a)}} d\gamma\\
=&\frac{2 b^a \mathsf{\Gamma} (a+\frac{1}{2})}{(2\pi)^\frac{1}{2} \mathsf{\Gamma}(a)} \left(b + \frac{x^2}{2}\right)^{-(a+\frac{1}{2})}u(x),
\end{align}
where $\mathsf{\Gamma}$ is defined as $\mathsf{\Gamma}(a) = \int_{0}^\infty t^{a-1} e^{-t} dt$.
More generally, all of the distributions represented by the GSM family have a corresponding rectified version represented by the R-GSM family (e.g. contaminated Normal and slash densities, symmetric stable and logistic, hyperbolic, etc.) \cite{andrews1974scale,palmer2006variational,palmer2005variational,lange1993normal,dempster1980iteratively,dempster1977maximum}.

\subsection{Relation to other Bayesian works}
In \cite{c1}, a modified Gaussian prior was considered for the NNLS problem. 
The authors used a Gaussian prior of arbitrary mean and variance and performed non-negative rectification using a `cut' function.
Their goal was to better represent \textit{non-sparse} signals by avoiding the selection of $\mu=0$, as we consider in our work.
Our R-GSM prior substantially differs from this work as we consider a \textit{mixture} of zero-location RG distributions for the prior, as opposed to a single Gaussian density with the `cut' rectification. 
Our design objective is to induce sparsity by using a hierarchical hyper-parameter $\vect{\gamma}$. 

In \cite{vila2014empirical}, a non-negative generalized approximate message passing (GAMP) approximation was proposed, using a Bernoulli non-negative Gaussian mixture prior of arbitrary location and scale parameters. 
This extends the prior given in \cite{c1} but uses a fixed number of mixture components e.g. $L=3$. 
The sparsity is enforced by using a Dirac delta function and an additional sparsity rate $\lambda$ that would `favor' the Dirac function and attenuate other mixture components simultaneously. 
The authors infer a bulk of parameters including the scale, location, and mixture weights as well as the sparsity rate simultaneously. 
Our R-SBL approach differs from \cite{vila2014empirical} as we only consider a single sparsity inducing hyper-parameter vector $\mathbf{\gamma}$, and our mixture components are strictly located at zero. 
Our approach simplifies the overall inference procedure and the problem formulation. 
We also consider an infinite number of mixture components as opposed to considering a fixed number of components.

Finally, we consider a more general class of priors than the existing methods since the R-GSM prior is based on an arbitrary mixing density $p(\mathbf{\gamma})$. 
As indicated in Section \ref{sec:example rgsm}, different selections of $p(\mathbf{\gamma})$ lead to more flexible and generalized priors for the sparse solution.

\section{Bayesian Inference with Scale Mixture Prior}
\label{sec:inference}
We detail the Type I and Type II methods for solving the S-NNLS problem with the R-GSM prior.
Though this paper is dedicated to Type II estimation because of its superior performance in sparse signal recovery problems
\cite{giri2015type,wipf2011latent}, we briefly introduce Type I in the following section for the sake of completeness.
\subsection{Type I estimation}
\label{sec:type1}
Using Type I to solve the S-NNLS problem translates into calculating the MAP
estimate of $\vectx$ given $\vecty$
\begin{align}\label{eq:type 1}
\argmin_{\vect{x}} \Vert \vecty - \vect\Phi \vectx \Vert_2^2 - \lambda\sum_{i=1}^M \ln p(x_i).
\end{align}
Some of the $\ell_0$ relaxation methods described in Section \ref{sec:intro} can be derived from a Type I perspective.
For instance, by choosing an exponential prior for $p(x_i)$, Eq.
\eqref{eq:type 1} reduces to the $\ell_1$ regularization approach in Eq. \eqref{nnl1} with the interpretation of $\lambda$
as being determined by the parameters of the prior and the noise variance. 
Similarly, by choosing a Gamma prior for $p(x_i)$, Eq. \eqref{eq:type 1} reduces to
\begin{align}
\argmin_{\vect{x}} \Vert \vecty - \vect\Phi \vectx \Vert_2^2 + \lambda\sum_{i=1}^M \ln \left(b+\frac{x_i^2}{2}\right),
\end{align}
which leads to the reweighted $\ell_2$ approach to the S-NNLS problem described in \cite{grady2008compressive,chartrand2008iteratively}. A unified Type I approach for the
R-GSM prior can be readily derived using the approaches discussed in \cite{giri2015type, palmer2006variational}.

\subsection{Type II estimation}
\label{sec:type2}
The Type II framework involves finding a ML estimate of $\vect{{\gamma}}$ using evidence maximization and approximating the posterior $p(\vectx|\vecty)$ by $p(\vectx|\vecty,\vect{{\gamma}}_{\text{ML}})$.
Then, appropriate point estimates and the solution $\vect{x}$ can be obtained. 
We refer to this approach as the \textit{rectified sparse Bayesian learning} (R-SBL).

Several strategies exist for estimating $\vectg$. 
The first strategy considers the problem of forming a ML estimate of $\vectg$ given $\vecty$  \cite{giri2015type,tipping2001sparse,gauvain1994maximum,bishop2006pattern}. 
In our case, $p(\vectg | \vecty)$ does not admit a closed form expression making this strategy difficult. 
The second strategy investigated here, aims to estimate $\vectg$ by using the EM algorithm \cite{giri2015type,c6,bishop2006pattern}. 
In the EM approach, we treat $(\vectx,\vecty,\vectg)$ as the complete data and $\vectx$ as the hidden variable. Utilizing the current estimate $\vectg^t$, where $t$ refers to the iteration index, the expectation step (E-step) involves finding the expectation of the log-likelihood, $Q(\vectg,\vectg^t)$ given by
\begin{align}
Q(\vectg,\vectg^t) =& E_{\vect{x} | \vecty; \vect{\gamma}^t} \left[ \ln p(\vect{y} | \vect{x}) + \ln p(\vect{x} | \vect{\gamma}) + \ln p(\vect{\gamma}) \right]\\
\dot{=}& \sum_{i=1}^M E_{\vect{x} | \vecty; \vect{\gamma}^t} \left[ -\frac{1}{2}\ln \gamma_i -\frac{x_i^2}{2\gamma_i}  + \ln p(\gamma_i) \right],
\end{align}
where $\dot{=}$ indicates that constant terms, and terms that do not depend on $\vect{\gamma}$ have been dropped since they do not affect the consequent M-step. 
For simplicity, we assume a non-informative prior on
$\vect{\gamma}$ \cite{tipping2001sparse}. In the M-step, we maximize $Q(\vectg,\vectg^t)$ with
respect to $\vectg$ by taking the derivative and setting it equal to zero,
which yields the update rule
\begin{align}
\label{eq:gamma update}
\gamma_i^{t+1} =E_{\vect{x}|\vect{y},\vect{\gamma}^t,\sigma^2}[x_i^2] \defeq \langle x_i^2 \rangle.
\end{align}
To compute $\langle x_i^2 \rangle$, we consider the multivariate posterior density $p(\vect{x}|\vect{y},\vect{\gamma},\sigma^2)$ which has the form (see Appendix \ref{sec:posterior})
\begin{align}\label{eq:posterior}
p(\vect{x}|\vect{y},\vect{\gamma}) =  c(\vect{y}) e^{-\dfrac{(\vect{x}-\vect{\mu})^T {\vect\Sigma}^{-1}(\vect{x}-\vect{\mu})}{2}} u(\vect{x}),
\end{align}
where $\vect{\mu}$ and $\vect\Sigma$ are given by \cite{tipping2001sparse,c6,c8}
\begin{align}
\label{mu}
\vect{\mu} = \vect{\Gamma}\vect\Phi^T(\sigma^2 \vect I+ \vect\Phi\vect\Gamma\vect\Phi^T)^{-1}\vect{y}~~~~~
\\
\vect\Sigma = \vect{\Gamma} - \vect{\Gamma}\vect\Phi^T(\sigma^2 \vect I+ \vect\Phi\vect{\Gamma}\vect\Phi^T)^{-1} \vect\Phi\vect{\Gamma} ,\label{sigma}
\end{align}
and $\vect{\Gamma}=\mbox{diag}(\vect{\gamma})$. 
The posterior in Eq. \eqref{eq:posterior} is known as a multivariate RG (or a multivariate truncated normal \cite{horrace2005some}). 
The normalizing constant $c(\vecty)$ does not admit a closed form expression. 
However, the M-step in Eq. \eqref{eq:gamma update} only requires the marginal density. 
Unfortunately, the marginals of a multivariate RG are not univariate-RG's and do not admit closed form expressions \cite{horrace2005some}, which also means no immediate expressions for the marginal moments.

However, we can approximate the first and the second moments $\langle x_i \rangle$ and $\langle x_i^2 \rangle$ of the multivariate RG posterior. 
In the following, we propose four different approaches for this purpose that offer a trade-off between computational complexity and theoretical accuracy. \\

\subsubsection{Markov Chain Monte Carlo EM (MCMC-EM)}\label{sec:mcmc}\hfill

Advances in numerical methods made it possible to sample from complex multivariate distributions \cite{robert2013monte, duane1987hybrid,pakman2014exact}.
Numerical methods are particularly useful when the first and second order statistics of a posterior density do not have a closed form expressions. 
In this case, the E-step can be performed by drawing samples using numerical Markov Chain Monte Carlo (MCMC) and then calculating the sample statistics. 
This approach is usually referred to as MCMC-EM \cite{andrieu2003introduction, sherman1999conditions}. 

First, we consider the Gibbs sampling approach in \cite{geman1984stochastic, li2015efficient}. 
We use hat notation to refer to the empirical estimates of various parameters (e.g. $\vect{\hat{\Sigma}}$, $\vect{\hat{\mu}}$).
We use the multivariate truncated normal (TN) definition in \cite{li2015efficient} and write
\begin{align}
&\mathsf{TN}(\vx; \vect{\hat{\mu}},\vect{\hat{\Sigma}},\vect{{R}},\vect \alpha_L,\vect \alpha_U) = \\
&\left(c_{tn} e^{-\dfrac{(\vect{x}-\vect{\hat{\mu}})^T {\vect{\hat{\Sigma}}}^{-1}(\vect{x}-\vect{\hat{\mu}})}{2}}\right) \mathbf{1}_{\vect\alpha_L \leq \vect{{R}} \vw \leq \vect\alpha_U} ,
\end{align}
where $\mathbf{1}_{(\cdot)}$ is the indicator function and $c_{tn}$ is the normalizing constant for the density. In the case of a multivariate rectified Gaussian, the truncation bounds are $\vect{\alpha}_L=\mathbf{0}$ and $\vect{\alpha}_U = \mathbf{\infty}$, and $\mathbf{{R}} = \mathbf{I}$. By introducing the transformation, $\vw = \vect{\hat{L}}\inv (\vectx-\vect{\hat{\mu}})$ where $\vect{\hat{L}}$ is the lower triangular Cholesky decomposition of $\vect{\hat{\Sigma}}$, it can be shown that $\vw$ is $\mathsf{TN}(\vw; \mathbf{0},\vect I,\vect{\hat{L}},\vect\alpha_L^*,\vect\alpha_U^*)$ with new truncation bounds $\vect \alpha_L^*=\vect \alpha_L -\vect{\hat{\mu}} = -\vect{\hat{\mu}}$ and $\vect \alpha_U^*=\vect \alpha_U -\vect{\hat{\mu}} = \vect{\infty}$.

The Gibbs sampler then proceeds by iteratively drawing samples from the conditional distribution
$p(w_i | \vecty,\vect{\hat{\gamma}},\sigma^2,\vect {w_{-i}})$, where $\vect {w_{-i}}$ refers to the vector containing all but the $i$th element of $\vw$. 
Given a set of samples drawn from $\vw$, we can obtain the samples for the original distribution of interest by inverting the transformation: $\lbrace \vectx^n \rbrace_{n=1}^N = \lbrace \vect{\hat{L}} \vw^n + \vect{\hat{\mu}} \rbrace_{n=1}^N$. 
Then, the first and second empirical moments can be calculated from the drawn samples using
\begin{align}
\langle {{x}_i} \rangle &\approx \frac{1}{N} \sum_{n=1}^N \left(x_i^n\right), \label{mcmc_fmoment} \\
\langle {{x}_i^2} \rangle &\approx \frac{1}{N} \sum_{n=1}^N \left(x_i^n\right)^2, \label{mcmc_secmoment}
\end{align}
and the EM can be iterated by updating $\hat{\gamma_i}^{t+1} = \langle {{x}_i^2} \rangle$.

After convergence, a point estimate for $\vect{x}$ is needed. The optimal estimator of $\vectx$ in the minimum mean-square-error (MMSE) sense is simply $\hat{\vectx}_{mean} = \langle {x}_i \rangle$. An alternative point estimate is to use $\hat{\vectx}_{mode}$ given by
\begin{align}
\hat{\vectx}_{mode} =& \argmax_{\vectx} p(\vectx | \vecty,\hat\vectg,\sigma^2)\\
=& \argmin_{\vx \geq 0} \Vert \vecty - \vect{\Phi}\vectx \Vert_2^2 + \lambda \sum_{i=1}^{M} \frac{x_i^2}{\hat{\gamma}_i}, \label{eq:mode estimator}
\end{align}
where Eq. \eqref{eq:mode estimator} can be solved by any NNLS solver. 
The estimate $\hat{\vectx}_{mode}$ could be a favorable point estimate because it chooses
the peak of ${p}(\vect{x}|\vecty,\hat{\vectg},\sigma^2)$, which may not be well-characterized by its mean.

For the sparse recovery problem at hand, we experienced very slow convergence with Gibbs sampling.
Convergence was particularly slow for higher problem dimensions and at larger cardinalities.
The latter was expected as a sparse solution is harder to recover in those cases.
Thus, we resorted to Hamiltonian Monte Carlo (HMC) which is designed specifically for target spaces constrained by linear or quadratic constraints \cite{pakman2014exact}. 
HMC improves the MCMC mixing performance by using the gradient information of the target distribution \cite{andrieu2003introduction}.

Despite use of the state of the art MCMC techniques, MCMC-EM might still converge to poor local minima solutions and result in sub-optimal performance \cite{neath2013convergence,bickel2008covariance,tong2014estimation}. 
Particularly, performance may be poorer for under-determined problems. 
Though MCMC-EM is not thoroughly investigated for the sparse recovery problem, here we list four major issues for consideration:
\begin{enumerate}[I.]
\item {Convergence:} MCMC-EM based algorithms can get stuck in a local minima depending on the problem dimensions and complexity of the search space. 
This is true even for well-posed problems \cite{sherman1999conditions,celeux1992stochastic}. 
In under-determined problems, the solution set for Eq. \eqref{intro_2} may contain many local minima and thus, a good MCMC-EM implementation should try to avoid local minima.
\item {Computational Limits:} Current MCMC sampling techniques are not optimal for drawing large sample sizes from high dimensional multivariate posterior densities. Therefore, the number of available samples is often limited by computational constraints \cite{robert2013monte, duane1987hybrid, pakman2014exact}.
\item {Quality of Parameter Estimates:} Since the MCMC samples are determined by random sampling at each iteration, the estimates of $\hat{\vect{\gamma}}$, $\hat{\vect{\mu}}$, and $\hat{\vect{\Sigma}}$ depend highly on the quality of the MCMC estimates $\hat{\vect{x}}$, which in turn affects the quality of next cycle of MCMC samples. This may lead the EM algorithm to converge to a sub-optimal solution.
\item {Structure of the Empirical $\hat{\vect{\Sigma}}$: } When $M$ is large and the dimensions of the empirical scale matrix are also large, $\hat{\vect{\Sigma}}$ may no longer be a good numerical estimate \cite{bickel2008covariance,fisher2011improved,ledoit2004well}. This issue could be exacerbated when the problem is inherently under-determined with $N<M$, and reveals itself as $\hat{\vect{\Sigma}}$ being close to singular. 
Therefore, regularization methods for $\hat{\vect{\Sigma}}$ are often used to alleviate this problem \cite{bickel2008covariance, tong2014estimation}.
\end{enumerate}
The scale matrix $\hat{\vect{\Sigma}}$ has direct control over the search space for MCMC and spurious off-diagonal values tend to increase
the number of local-minima. Therefore, to address the issues listed above, we incorporated ideas from prior work to regularize the estimates of $\hat{\vect{\Sigma}}$:
\begin{itemize}
\item As in \cite{bickel2008covariance,tong2014estimation}, we assume that $\hat{\vect{\Sigma}}$ is sparse and we prune its off-diagonal entries when they drop below a certain threshold $T_p$. This prevents the spurious off-diagonal values in $\hat{\vect{\Sigma}}$ from affecting the next cycle of MCMC samples and improves future estimates of $\hat{\vect{\gamma}}$.
\item We incorporate the shrinkage estimation idea presented in \cite{tong2014estimation,fisher2011improved} and regularize $\hat{\vect{\Sigma}}$ as a convex sum of the empirical $\hat{\vect{\Sigma}}$ and a target matrix $\vect{T}$ such that, $\hat{\vect{\Sigma}} = \lambda \hat{\vect{\Sigma}} + (1-\lambda)\vect{T}$. A simple selection for $\vect{T}$ is the matrix $\hat{\vect{\Sigma}}_{\beta}$, which is equal to the original $\hat{\vect{\Sigma}}$ with diagonal elements scaled by a factor $\beta$. 
Though this approach does not guarantee convergence to a global minimum and the solution could still be a local minima or a saddle point solution, we empirically observed better recovery performance.
\\
\end{itemize}
\subsubsection{Linear minimum mean-square-error (LMMSE)}\hfill

The LMMSE estimation approach is motivated by the complexity of the MCMC-EM approach.
Examining the parameters being computed, one can interpret them as finding the MMSE estimate of $\vectx$ and the associated MSE. 
This motivates replacing the MMSE estimate by the simple LMMSE estimate of $\vectx$.
The affine LMMSE estimate for $\vect{x}$ is
\begin{align}\label{eq:lmmse}
\vect{\hat{x}} = \vect{\mu_{x}} + \vect{R_{x}}\vect{\Phi}^T( \vect{\Phi}\vect{R_{x}} \vect{\Phi}^T + \sigma^2\vect{I} )^{-1}(\vect{y}-\vect{\Phi}\vect{\mu_x}) ,
\end{align}
where $\vect{R_{x}}$ is the covariance matrix of $\vectx$ (a diagonal matrix).
The estimation error covariance matrix is given by \cite{kailath2000linear}
\begin{align}\label{eq:lmmse_esterr}
\vect{R_{e}} =  \vect{R_{x}} -  \vect{R_{x}} \vect{\Phi}^T( \vect{\Phi}\vect{R_{x}} \vect{\Phi}^T + \sigma^2\vect{I} )^{-1}\vect{\Phi}\vect{R_{x}}.
\end{align}
To elaborate, in the E-step where $\vectg$ is fixed at $\vectg^t,$ the entries of $\vect{x}$ are independent, and the prior mean and the prior covariance will be equal to the mean and variance of the independent univariate RG distributions with $p(x_i|\gamma_i) = \mathcal{N}^R(0,\gamma_i)$. 
The mean of a univariate rectified Gaussian density with zero location parameter is given by \cite{miskin2000ensemble}
\begin{align}
\label{eq:mean_lmmse}
{\mu_{x,i}}= \sqrt{\dfrac{2\gamma_{i}}{\pi}},
\end{align}
and the variances which are the diagonal entries of the diagonal matrix $\vect{R_{x}}$ are given by
\begin{align}
\label{eq:var_lmmse}
R_{x,ii} = \gamma_i \left( 1 -2/\pi \right).
\end{align}
Using the values of $\vect{\mu_{x}}$ and $\vect{R_{x}}$ from Eq. \eqref{eq:mean_lmmse} and Eq. \eqref{eq:var_lmmse} in Eq. \eqref{eq:lmmse} we obtain the
LMMSE point estimate for the solution vector. Similarly, the update for $\vect{\gamma}$ (M-step) is given by
\begin{align}
\label{eq:gamma_lmmse}
{\gamma_i} = {\hat{x}_i}^2 + {R_{e,ii}}.
\end{align}
This is sufficient to implement the EM algorithm. Upon convergence, the mean point estimate is simply $\hat{\vectx}_{mean} = \vect{\hat{x}}$, and the mode point estimate can be obtained by utilizing the converged values ${\gamma_i}$ in Eq. \eqref{eq:mode estimator}.
~
 \\
\subsubsection{Generalized approximate message passing (GAMP)}\hfill

In this section, we present an EM implementation using the generalized approximate message passing (GAMP) algorithm \cite{al2018gamp,rangan2011generalized}. 
A different GAMP based approach was used in \cite{vila2014empirical}, which uses an i.i.d. Bernoulli non-negative Gaussian mixture prior with a fixed mixture order that is independent of $M$. 
To overcome the convergence issues with the type of GAMP algorithm in \cite{vila2014empirical} e.g. when a non-i.i.d. design matrix $\boldsymbol{\Phi}$ is used \cite{rangan2014convergence,caltagirone2014convergence,Schniter_conv}, we incorporate the damping technique in \cite{al2018gamp,Schniter_conv} into the proposed R-SBL GAMP algorithm.

GAMP is a low complexity iterative inference algorithm. 
The low complexity is achieved by applying quadratic and Taylor series approximations to loopy belief propagation. 
GAMP can approximate the MMSE estimate when used in the sum-product version, or can approximate the MAP estimate when used in the max-sum version.
The sum-product version computes the mean and variance of the approximate marginal posteriors on $x_i$ which are given by
\begin{flalign}
&p(x_i \mathcal{\vert} r_i;\tau_{r_i}) \propto p(x_i) \mathcal{N} (x_i ; r_i,\tau_{r_i}) \label{marginal},
\end{flalign}
where $r_i$ approximates an AWGN corrupted version of the true $x_i$ as
\begin{flalign}
&r_i \approx x_i+\bar{r}_i \label{AWGN_app}\\
&\bar{r}_i \sim \mathcal{N}(0,\tau_{r_i}).
\end{flalign}

In the large system limit and when the design matrix $\boldsymbol{\Phi}$ is i.i.d sub-Gaussian, the approximation in Eq. \eqref{AWGN_app} was shown to be exact \cite{rangan2011generalized,javanmard2013state}. 
Therefore, in the sum-product version of GAMP, the estimate $\hat{x}_i$ in Eq. \eqref{x_sum_product} corresponds to the MMSE estimate of $x_i$ given $r_i$, and similarly the conditional variance of $x_i$ given $r_i$ is defined in Eq. 
\eqref{var_sum_product}.
\begin{flalign}
\hat{x}_i&= \mathbb{E}\{ x_i \mathcal{\vert} r_i;\tau_{r_i} \} \label{x_sum_product} \\
\tau_{x_i}&=\text{var} \{ x_i \mathcal{\vert} r_i;\tau_{r_i} \}\label{var_sum_product}.
\end{flalign}

In the max-sum version of GAMP, the MAP estimate $\hat{x}_i$ given $r_i$ is obtained in Eq. \eqref{x_max_sum} using the proximal operator defined in Eq. \eqref{prox_op}, while  $\tau_{x_i}$ given in Eq. \eqref{var_max_sum} corresponds to the sensitivity of the proximal thresholding.
\begin{flalign}
\hat{x}_i&= \text{prox}_{-\ln p(x_i)} (r_i;\tau_{r_i})\label{x_max_sum}\\
\tau_{x_i}&=\tau_r {\text{prox\ensuremath{'}}}_{-\ln p(x_i)} (r_i;\tau_{r_i}) \label{var_max_sum}\\
\text{prox}_{f} (\hat{a},\tau^a) &\triangleq \argmin_{x \in \boldsymbol{R}} f(x)+ \frac{1}{2 \tau^a} |x-\hat{a} |^2.\label{prox_op}
\end{flalign}

When implementing the EM algorithm, the approximate posterior computed by the sum-product GAMP can be used to efficiently approximate the E-step \cite{vila2013expectation}. 
Moreover, in the case of max-sum GAMP, in the large system limit and under i.i.d sub-Gaussian $\boldsymbol{\Phi}$ an extra step can be added as in \cite{vila2014empirical} to compute the marginal distributions using Eq. \eqref{marginal}. 
These marginals then can be used to approximate the E-step.
For the rectified Gaussian scale mixture prior $p(\boldsymbol{x} \mathcal{\vert} \boldsymbol{\gamma})$ the details of finding $\hat{x}_i$ and $\tau_{x_i}$ estimates in both the sum-product and max-sum cases are shown in Appendix \ref{sec:gamp}.

\begin{table}[t!]
\centering
\begin{tabular}{|l r|}
\hline
Initialization & \\
$\boldsymbol{S} \leftarrow |\boldsymbol{\Phi}|^2$ (component wise magnitude squared)  & \\
Initialize $\dot{\boldsymbol{\tau}}^0_x, {\boldsymbol{\gamma}}^0>0$ & \\
$\dot{\boldsymbol{s}}^0,\dot{\boldsymbol{x}}^0 \leftarrow \boldsymbol{0}$  & \\
\hline
for $i=1,2,....,I_\text{max}$ & \\
\ \ \ Initialize $\boldsymbol{\tau}^{1}_x \leftarrow \dot{\boldsymbol{\tau}}^{i-1}_x , \hat{\boldsymbol{x}}^{1} \leftarrow \dot{\boldsymbol{x}}^{i-1}, {\boldsymbol{s}}^{1} \leftarrow \dot{\boldsymbol{s}}^{i-1}$  &\\
\ \ \ // E-Step Approximation & \\
\ \ \ for $k=1,2,....,K_\text{max}$ & \\
\ \ \ $\ \ \ 1/\boldsymbol{\tau}_p^k \leftarrow \boldsymbol{S} \boldsymbol{\tau}_x^k$ & \\
\ \ \ $\ \ \ \boldsymbol{p}^k \leftarrow \boldsymbol{s}^{k-1} + \boldsymbol{\tau}_p^k \boldsymbol{\Phi} \hat{\boldsymbol{x}}^k$ & \\
\ \ \ $\ \ \ \boldsymbol{\tau}_s^k \leftarrow \frac{\sigma^{-2} \boldsymbol{\tau}_p^k}{\sigma^{-2} + \boldsymbol{\tau}_p^k}$ & \\
\ \ \ $\ \ \ \boldsymbol{s}^k \leftarrow (1-\theta_s)\boldsymbol{s}^{k-1} + \theta_s (\boldsymbol{p}^k/\boldsymbol{\tau}_p^k-\boldsymbol{y})/(\sigma^2+1/\boldsymbol{\tau}_p^k)$ & \\
\ \ \ $\ \ \ 1/\boldsymbol{\tau}_r^k \leftarrow \boldsymbol{S}^\top \boldsymbol{\tau}_s^k$ & \\
\ \ \ $\ \ \ \boldsymbol{r}^k \leftarrow \hat{\boldsymbol{x}}^k - \boldsymbol{\tau}_r^k \boldsymbol{\Phi}^\top \boldsymbol{s}^k$ & \\
\ \ \ \ \ \ if MaxSum then &\\
\ \ \ $\ \ \ \ \ \ \boldsymbol{\tau}_x^{k+1} \leftarrow \boldsymbol{\nu}^k$  & \\
\ \ \ $\ \ \ \ \ \ \hat{\boldsymbol{x}}^{k+1} \leftarrow \boldsymbol{\eta}^k u(\boldsymbol{r}^k)\ \ \ $ & \\
\ \ \ \ \ \ else &\\
\ \ \ $\ \ \ \ \ \ \boldsymbol{\tau}_x^{k+1} \leftarrow \boldsymbol{\nu}^k g(\frac{\boldsymbol{\eta}^k}{\boldsymbol{\nu}^k})$  & \\
\ \ \ $\ \ \ \ \ \ \hat{\boldsymbol{x}}^{k+1} \leftarrow \boldsymbol{\eta}^k + \sqrt{\boldsymbol{\nu}^k} h(\frac{\boldsymbol{\eta}^k}{\boldsymbol{\nu}^k})$ & \\
\ \ \ \ \ \ end if &\\
\ \ \ \ \ \ if $\|\hat{\boldsymbol{x}}^{k+1}-\hat{\boldsymbol{x}}^k\|^2 / \|\hat{\boldsymbol{x}}^{k+1}\|^2 < \epsilon_\text{gamp}$ , break & \\
\ \ \ end for \%end of k loop & \\
\ \ \ $\dot{\boldsymbol{s}}^i \leftarrow \boldsymbol{s}^{k}$ &\\
\ \ \ if MaxSum &\\
\ \ \ \ \ \ $\dot{\boldsymbol{x}}^i \leftarrow \boldsymbol{\eta}^{k+1} + \sqrt{\boldsymbol{\nu}^{k+1}} h(\frac{\boldsymbol{\eta}^{k+1}}{\boldsymbol{\nu}^{k+1}})$ , $\dot{\boldsymbol{\tau}}^i_x \leftarrow \boldsymbol{\nu}^{k+1} g(\frac{\boldsymbol{\eta}^{k+1}}{\boldsymbol{\nu}^{k+1}})$ &\\
\ \ \ else &\\
\ \ \ \ \ \ $\dot{\boldsymbol{x}}^i \leftarrow \hat{\boldsymbol{x}}^{k+1}$ , $\dot{\boldsymbol{\tau}}^i_x \leftarrow \boldsymbol{\tau}^{k+1}_{x}$ &\\
\ \ \ end if &\\
\ \ \ // M-Step & \\
\ \ \ $\boldsymbol{\gamma}^{i+1} \leftarrow |\dot{\boldsymbol{x}}^i|^2+\dot{\boldsymbol{\tau}}^i_x$ &\\
\ \ \ if $\|\dot{\boldsymbol{x}}^{i}-\dot{\boldsymbol{x}}^{i-1}\|^2 / \|\dot{\boldsymbol{x}}^{i}\|^2 < \epsilon_\text{em}$ , break & \\
end for \%end of i loop & \\
\hline
\end{tabular}
\caption{R-SBL GAMP Algorithm}
\label{tab:GAMP-RSBL}
\end{table}
Upon convergence of the GAMP algorithm, the approximate E-step of the EM algorithm is complete and we can evaluate the M-step in Eq. \eqref{eq:gamma update} as
\begin{flalign}
&\langle x_i^2 \rangle = \int_{x_i} x^2_i p(x \mathcal{\vert} r_i;\tau_{r_i}) =  \hat{x}^2_i + \tau_{x_i}.
\end{flalign}

The EM-based R-SBL GAMP algorithm is summarized in Table \ref{tab:GAMP-RSBL}. 
Here, the steps used by the GAMP algorithm to evaluate $\boldsymbol{s}$ and $\boldsymbol{\tau}_s$ are the same for both sum-product and max-sum versions (for AWGN case) \cite{rangan2011generalized}.
In Table \ref{tab:GAMP-RSBL}, all mathematical operations are element wise. 
$K_\text{max}$ is the maximum allowed number of GAMP iterations, $\epsilon_\text{gamp}$ is the GAMP tolerance parameter, $I_\text{max}$ is the maximum number of EM iterations, and $\epsilon_\text{em}$ is the EM tolerance parameter. Also, ${\theta_s}\in (0,1]$ is the damping factor which can be selected according to the empirical criteria in \cite{al2018gamp}, and $\boldsymbol{\eta}$, $\boldsymbol{\nu}$, $h(.)$, and $g(.)$ are defined in Appendix \ref{sec:gamp}.
~\\
\vspace{-0.01em}
\subsubsection{Diagonal approximation (DA)}\hfill

We know \textit{a-priori} that the posterior in Eq. \eqref{eq:posterior} does not admit a closed form expression.
However, to implement the EM algorithm we only need the marginal moments of the posterior.
We first note that, if the scale matrix $\vect{\Sigma}$ is diagonal then we could evaluate the normalizing constant $c(\vect{y})$ in closed form since the multivariate RG posterior can be written as a product of univariate marginals (see Appendix \ref{sec:posterior}).

In the diagonal approximation (DA) approach, we resort to approximating the posterior in Eq. \eqref{eq:posterior} with a suitable posterior density ${p}(\vect{x}|\vect{y},\vect{\gamma}) \approx \tilde{p}(\vect{x}|\vect{y},\vect{\gamma})$, which could be written as a product of independent marginal densities
i.e. $\tilde{p}(x_i|\vect{y},\vect{\gamma})$. 
This approximate posterior density is derived in Appendix \ref{sec:posterior} as
\begin{align}\label{appost_51}
\tilde{p}(\vectx|\vecty,\vectg)  =&  \prod_{i=1}^{M} \tilde{p}(x_i|\vect{y},\vect{\gamma})\\
=& \prod_{i=1}^{M}  \sqrt{\dfrac{2}{\pi\Sigma_{ii}}} \dfrac{e^{-\dfrac{(x_i-\mu_i)^2}{2\Sigma_{ii}}}u(x_i)}{
 \erfc\left(-\dfrac{\mu_i}{\sqrt{2{{\Sigma}_{ii}}}}\right)},
\label{appost_5}
\end{align}
where $\mu_{i}$ is the $i$th element of $\vect{\mu}$ and $\Sigma_{ii}$ is the $i$th diagonal element of $\vect{\Sigma}$ obtained using Eqs. \eqref{mu} and \eqref{sigma}.
The marginal $\tilde{p}(x_i|\vect{y},\vect{\gamma})$ in Eq. \eqref{appost_51} is the univariate RG density defined in Eq. \eqref{rgg_density}, where $\tilde{p}(x_i|\vect{y},\vect{\gamma})= \mathcal{N}^{R}(x_i;\mu_i,\Sigma_{ii})$. 
Then, the univariate RG marginals are well-characterized by their first and second moments given in \cite{miskin2000ensemble}, with the first moment given as
\begin{align}
\label{eq:mean diag}
\langle x_i \rangle = \mu_i+\sqrt{\dfrac{2\Sigma_{ii}}{\pi}}\dfrac{e^{-\frac{\mu_i^2}{2\Sigma_{ii}}}}{\erfc \left(-\frac{\mu_i}{\sqrt{2\Sigma_{ii}}}\right)},
\end{align}
and the second moment given as
\begin{align}
\label{eq:approx diag}
\langle x_i^2 \rangle = \mu_i^2 + \Sigma_{ii} +\mu_i \sqrt{\dfrac{\Sigma_{ii}}{\pi}}\dfrac{e^{-\frac{\mu_i^2}{2\Sigma_{ii}}}}{\erfc \left(-\dfrac{\mu_i}{\sqrt{2\Sigma_{ii}}}\right)}.
\end{align}
Note that the moments of $\tilde{p}(x_i|\vect{y},\vect{\gamma})$ are approximations to the moments of the true marginals which do not admit closed form.
However, we can perform EM using the approximate moments to approximate the true solution.
EM can be carried out by setting  $\gamma_i^{t+1} = \langle x_i^2 \rangle$ and
iterating over $t$. 
After convergence of $\gamma_i$s, the mean point estimate is obtained as $\hat{\vectx}_{mean} = \langle x_i \rangle$. The mode point estimate $\hat{\vectx}_{mode}$ can be calculated by using converged values of $\gamma_i$s in Eq. \eqref{eq:mode estimator}. 
\begin{figure}[t!]
\centering
\includegraphics[clip,scale =0.45]{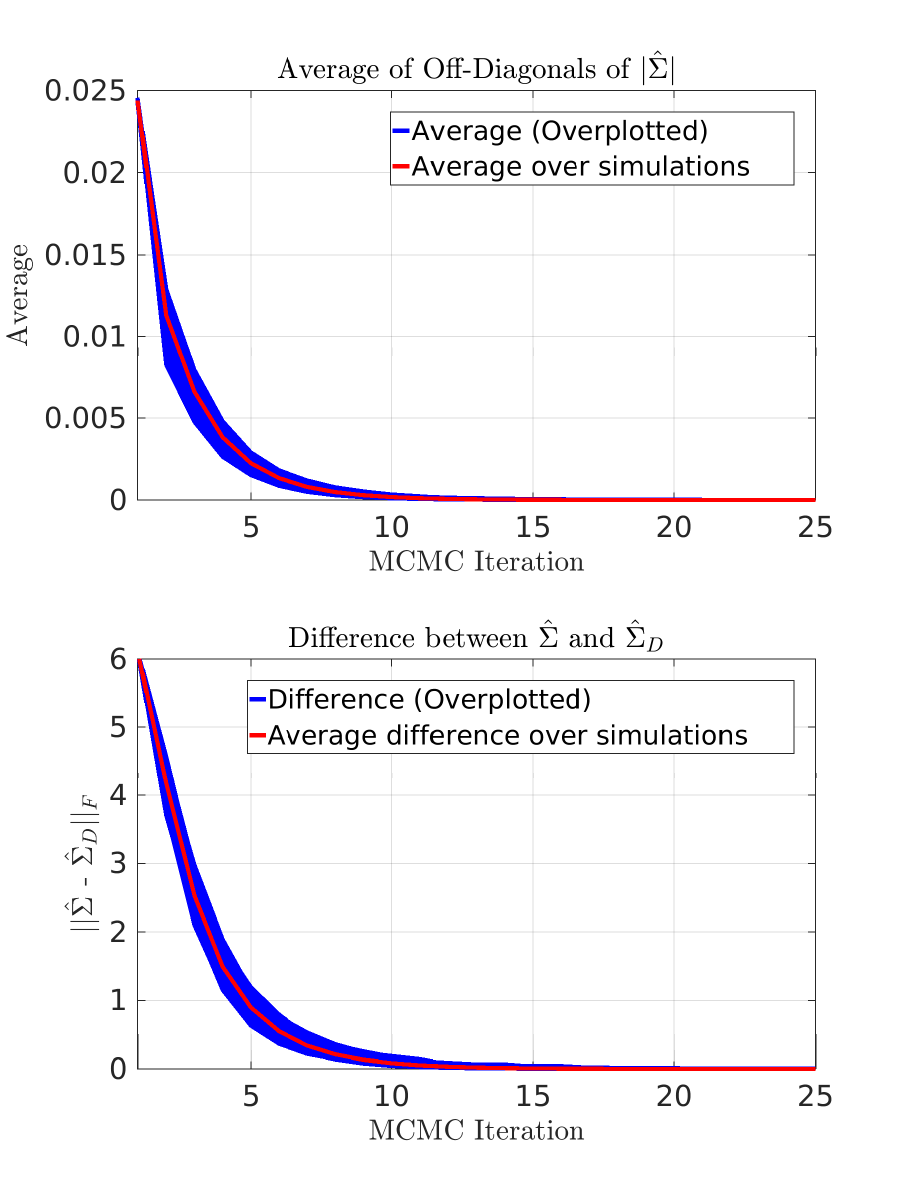}
\caption{\textit{Top}: Empirical observations for the structure of $\vect{\Sigma}$. We performed S-NNLS recovery using MCMC-EM (without regularizing the estimates of $\vect{\Sigma}$) and monitored the average value of off-diagonals for $|\vect{\Sigma}|$. 
We simulated for 1,000 runs and overplotted the results (blue lines). 
The average of average off-diagonals for $|\vect{\Sigma}|$ over 1,000 results is shown with the red line.  
The exponentially decreasing behavior suggests that the off-diagonal magnitudes of $\vect{\Sigma}$ decrease over MCMC iterations, indicating that true $\vect{\Sigma}$ is approaching to a diagonal form. 
\textit{Bottom}: The distance between the true $\vect{\Sigma}$ and a diagonal matrix formed by its diagonal entries $\vect{\Sigma}_D$. This suggests that the true $\vect{\Sigma}$ approaches to a diagonal form over MCMC iterations.}
\label{fig_f5_sigma}
\end{figure}

If the diagonal elements of $\vect{\Sigma}$ are large valued or become large over EM iterations as compared to the off-diagonals, then DA is expected to work well. 
Note that assuming a diagonal $\vect{\Sigma}$ was also motivated by prior work \cite{bickel2008covariance,fisher2011improved,ledoit2004well,horrace2005ranking,magdon2010approximating}
for various applications. In this work, we empirically report that DA has very good sparse recovery performance and has low complexity.

To further support the DA approximation, we present empirical findings regarding the structure of $\vect{\Sigma}$.
We performed sparse recovery simulations using Eq. \eqref{intro_1} with the MCMC-EM approach as the ground truth (without regularizing the MCMC estimates of $\hat{\vect{\Sigma}}$).
We assumed that $\vect{x}$ was of size 200 with 10 non-zero elements drawn from $\mathcal{N}^R({0}, 1)$. The dictionary $\vect{\Phi} \in \mathbb{R}^{50\times 200}$ columns were normally distributed $\vect{\Phi} \sim \mathcal{N}(0, \vect I)$. 
We solved this problem for 1,000 simulations and overlay plots of the average absolute value of the off-diagonals of $\hat{\vect{\Sigma}}$ as a function of MCMC-EM iteration in the first row of Fig. \ref{fig_f5_sigma} (blue lines).

We see that the average off-diagonal elements of $|\hat{\vect{\Sigma}}|$ exponentially approach $0$ as a function of MCMC-EM iteration.
The average of this behavior over 1,000 simulations (red line) has a final value of $10^{-4}$ after 10 iterations. 
This indicates that the off-diagonals of $\hat{\vect{\Sigma}}$ of the true posterior (with MCMC sampling) approach zero. 
Moreover, in the second row of Fig. \ref{fig_f5_sigma} we overlay plots of the Frobenius norm of the difference between $\hat{\vect{\Sigma}}$ and $\hat{\vect{\Sigma}}_D$, where $\hat{\vect{\Sigma}}_D$ is the diagonal matrix consisting of diagonal elements from $\hat{\vect{\Sigma}}$. 
This shows that as MCMC-EM converges $\hat{\vect{\Sigma}}$ approaches a diagonal form. 

These results suggest that, if there is flexibility in choosing the dictionary $\vect{\Phi}$ as in compressed sensing, then proper choice of $\vect{\Phi}$ can lead to the DA approach producing high quality approximate marginals $\tilde{p}(x_i|\vect{y},\vect{\gamma})$ that are close to the true marginals.

\vspace{-0.01em}
\subsection{Computational complexity of proposed methods}\label{time_comp}
For computational comparisons, we assume that $N\leq M$. 
Under this assumption, the time complexity of the DA algorithm is $\mathcal{O}(N^2M)$ per EM iteration.
This complexity is similar to the original SBL algorithm in \cite{wipf2007empirical,c6} and is due to the computationally intensive matrix inversion step $(\sigma^2 \vect I+ \vect\Phi\vect{\Gamma}\vect\Phi^T)^{-1}$ given in Eq. \eqref{mu}.
Time complexity of the LMMSE algorithm is also $\mathcal{O}(N^2M)$ per EM iteration. 
This complexity is determined from a similar matrix inversion step $( \vect{\Phi}\vect{R_{x}} \vect{\Phi}^T + \sigma^2\vect{I} )^{-1}$  in Eq. \eqref{eq:lmmse_esterr} (note that $\vect{R_{x}}$ is diagonal).
The GAMP algorithm bypasses the computationally intensive matrix inversion and the resulting complexity is $\mathcal{O}(NM)$ time \cite{al2018gamp}. This is linear in both problem dimensions and significantly faster than the both the DA and LMMSE methods.
For the MCMC-EM algorithm, the actual computational cost is determined by the random Hamiltonian MCMC sampling, which is explained in more detail in \cite{pakman2014exact}. 

\vspace{-0.01em}
\section{Experiment Design}
\label{sec:setup}
\begin{figure*}[h!]
\centering
\includegraphics[scale =.3905]{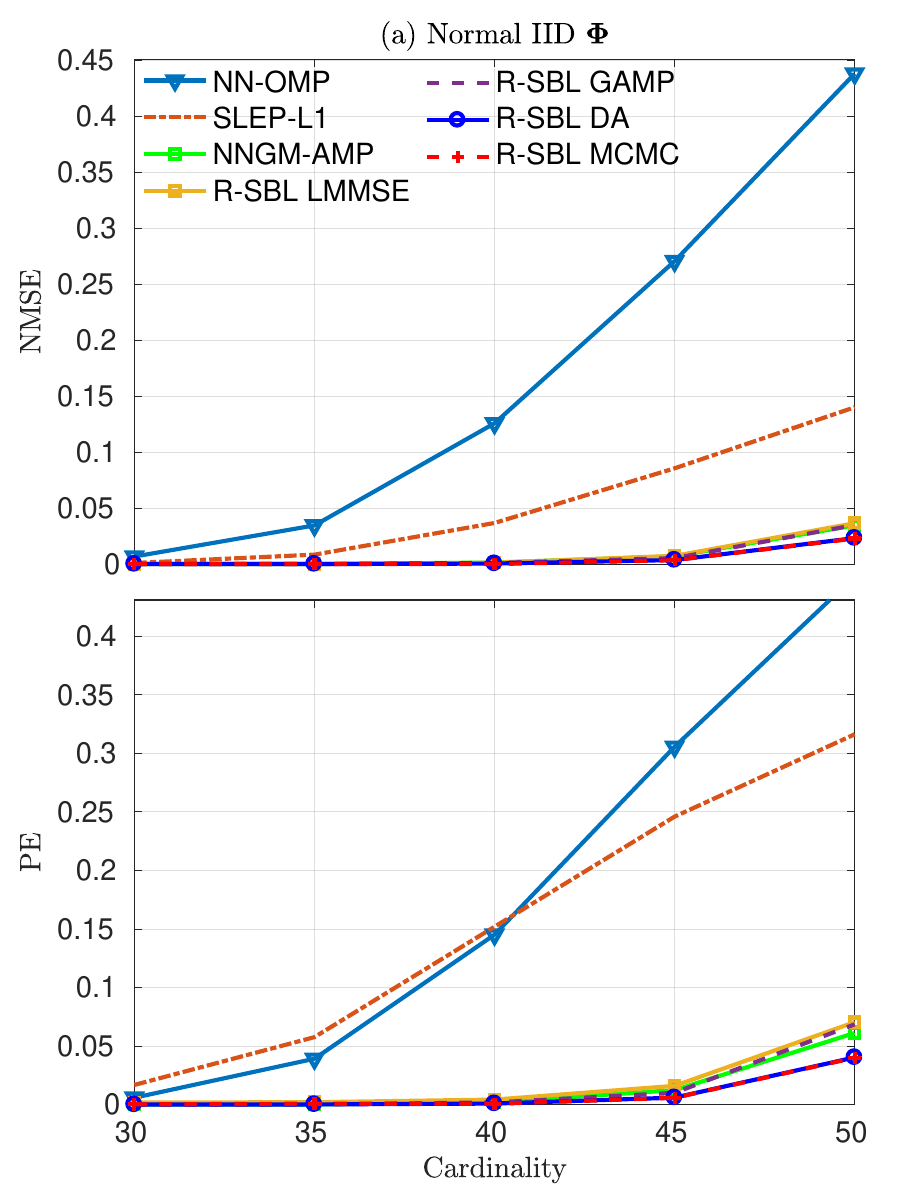}
\includegraphics[scale =.3905]{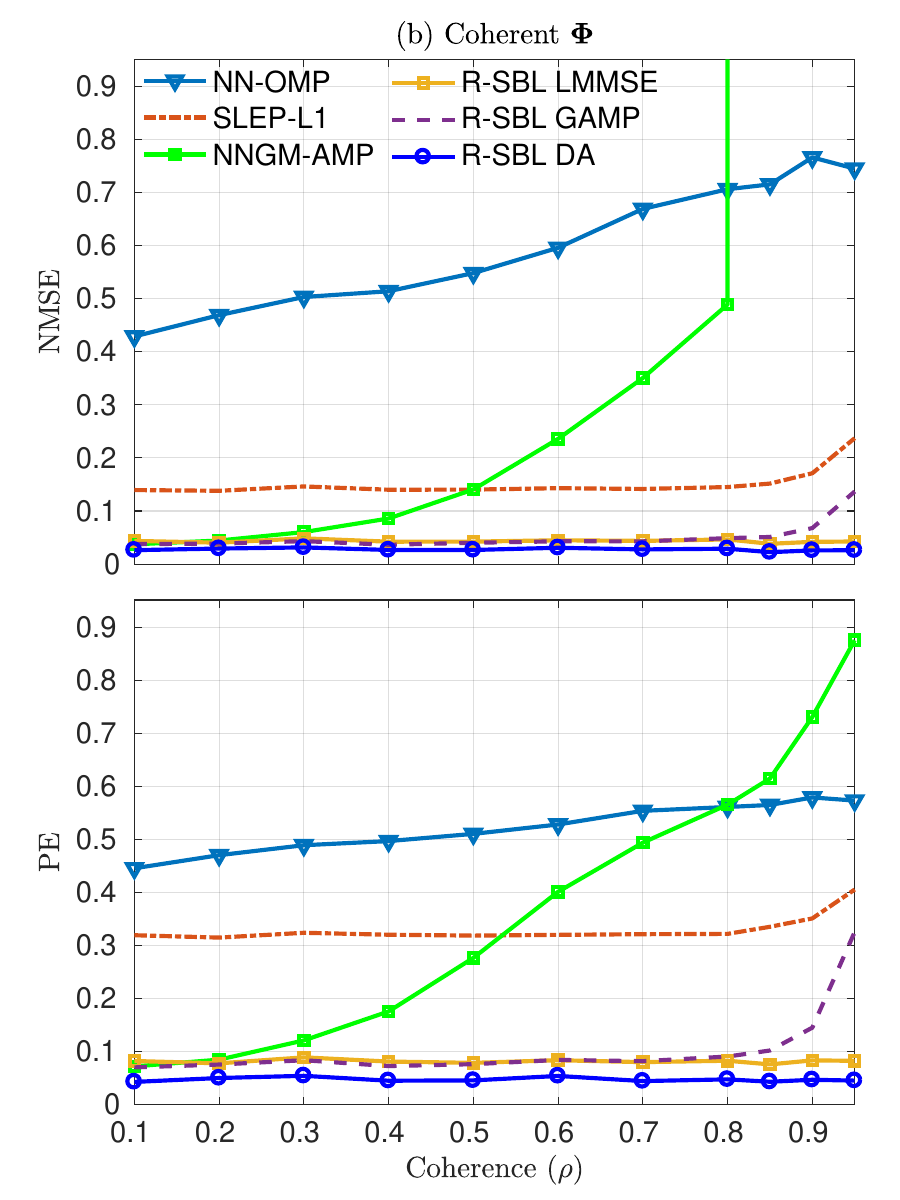}
\includegraphics[scale =.3905]{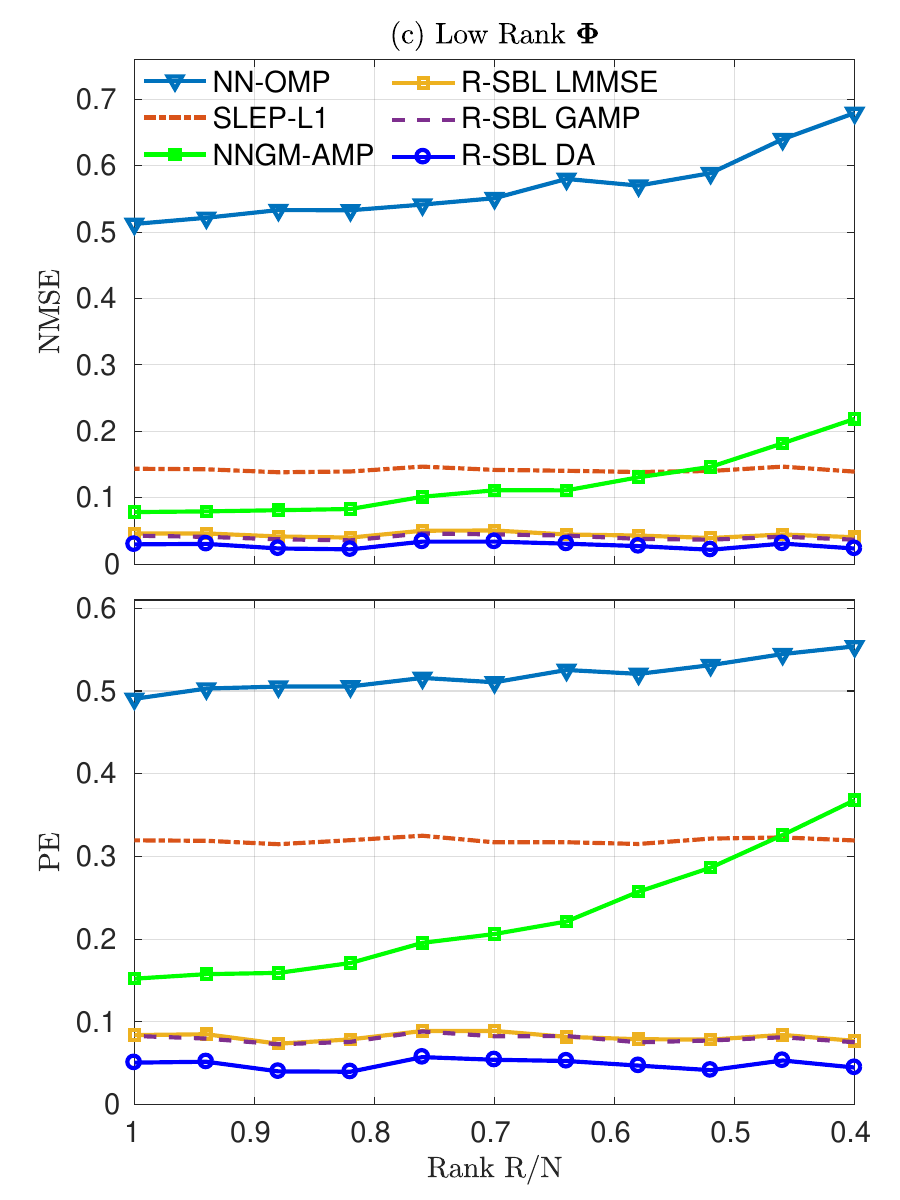}
\caption{Sparse recovery performances (NMSE and PE) of the R-SBL variants and the baseline S-NNLS solvers for various $\vect{\Phi}$.
In (a) the dictionary elements were i.i.d Normal and the sparse recovery results are shown for cardinalities $K=30$ to $K=50$. 
R-SBL DA achieves the best recovery performance. 
R-SBL LMMSE and GAMP are similar to NNGM-AMP and are much better than SLEP-$\ell_1$ and NN-OMP.
In (b) the dictionary columns are coherent with the coherence degree $\rho$ indicated in the x-axis. 
R-SBL variants are extremely robust to increasing coherence and result in a very small NMSE and PE across all $\rho$ values. 
NNGM-AMP breaks down after $\rho=0.2$ with deteriorating performance with increasing $\rho$ and SLEP-$\ell_1$ is better than NNGM-AMP after $\rho=0.5$.
In (c) the dictionary is rank-deficient with rank-ratio $R/N$ indicated in the x-axis. 
R-SBL variants are superior to baseline methods across all $R/N$ values.}
\label{fig_f1_norm}
\end{figure*}

In this section we provide the layout of our numerical experiments.
We provide extensive comparisons between the proposed R-SBL variants LMMSE, GAMP, MCMC and DA and the baseline S-NNLS solvers, including NNGM-AMP \cite{vila2014empirical}, SLEP-$\ell_1$ \cite{liu2009slep}, and NN-OMP \cite{bruckstein2008uniqueness}. 
In all of the experiments below, we generate sparse vectors $\vect{x^{gen}} \in \mathbb{R}^{400}_+$, such that $||\vect{x^{gen}}||_0 = K$, and random dictionaries $\vect{\Phi} \in \mathbb{R}^{100 \times 400}$. 
We normalize the columns of $\vect{\Phi}$ by $1/\sqrt{N}$ \cite{candes2006stable}. 
For a fixed $\vect{\Phi}$ and $\vectx^{gen}$, we compute the measurements $\vecty = \vect{\Phi}\vectx^{gen}$ and use the baseline algorithms and the proposed R-SBL variants to approximate $\vectx^{gen}$.

In the first set of experiments, we simulate a `noiseless' recovery scenario, where the noise variance is set as $\sigma^2 = 10^{-6}$, the non-zero entries of the solution vector are drawn from a rectified Gaussian density $\mathcal{N}^R({0}, 1)$ and the dictionary columns are i.i.d. Normal distributed $\vect{\Phi} \sim \mathcal{N}({0}, \vect I)$. 
We experiment with cardinalities $K=\{10,20,30,35,40,45,50\}$.

In the second set, we construct various dictionary types to analyze the robustness of our R-SBL method and the baseline solvers for the S-NNLS problem. 
The dictionary types considered here are not necessarily i.i.d. Gaussian and are similar to the ones used in \cite{al2018gamp,vila2015adaptive}. 
These dictionaries can be low-rank, coherent, ill-posed, and non-negative as detailed below:

\begin{enumerate}[A.]
\item \textit{Coherent dictionaries:} We introduce coherence among the columns of an original dictionary $\vect{\Phi}= \mathcal{N}(0, \vect I)$ and report recovery performances for a fixed $K=50$. 
This was done by multiplying $\vect{\Phi}$ with a coherence matrix $\vect{C}$ to obtain a new dictionary $\vect{\Phi}_c$ with coherent columns. 
Here, $\vect{C}$ is the Cholesky factor of the Toeplitz($\rho$) matrix with a coherence parameter $\rho$. 
We experiment with different coherence values by selecting $\rho=\{0.1, 0.2,...,0.80, 0.85, 0.90 ,0.95\}$.
\item \textit{Low-rank dictionaries:} We construct rank-deficient dictionaries such that $\vect{\Phi} = \vect{A}\vect{B}$, where $\vect{A} \in \mathbb{R}^{N \times R}$, $\vect{B} \in \mathbb{R}^{R \times M}$ and $R<N$. 
The entries of $\vect{A}$ and $\vect{B}$ are i.i.d. Normal. 
The rank ratio $R/N$ is considered as a measure of rank deficiency, where smaller values indicate more deviation from an i.i.d. dictionary. 
We experiment with $R/N=\{1, 0.95, ...,0.4\}$ and report recovery performances for a fixed $K=50$.
\item \textit{Ill-conditioned dictionaries:} We experiment with ill-conditioned dictionaries with a condition number $\kappa>1$.  
For a fixed $\kappa$, the dictionary is constructed as $\vect{\Phi} = \vect{U}\vect{S}\vect{V}^T$. 
Here, $\vect{U}$ and $\vect{V}$ contain the left and right singular vectors of an i.i.d. Gaussian matrix, and $\vect{S}$ is a diagonal matrix containing the eigenvalues. 
We decay the elements of $\vect{S}$ with  $\vect{S}_{i+1,i+1} = \kappa^{-1/{(N-1)}} \vect{S}_{i,i}$ for $i=1,2,...,N-1$. 
The value of $\kappa$ measures the deviation from an i.i.d. Gaussian dictionary, with larger $\kappa$ values indicate more deviation.
We experiment using the condition numbers $\kappa=\{8, 10,...,28\}$.
\item \textit{Non-negative dictionaries:} Non-negative dictionaries are used in sparse recovery applications such as sparse NMF \cite{peharz2012sparse} and NN K-SVD \cite{aharon2005k}, where a positive mapping is required on the solution vector. 
We construct non-negative dictionaries $\vect{\Phi}$ with columns that are drawn according to $\vect{\Phi} \sim RG({0}, \vect I)$. 
We experiment with cardinalities $K=\{10,20,30,35,40,45 ,50\}$.
\end{enumerate}

In the third set of experiments, we set the noise variance $\sigma^2$ for $\vect{v}$ such that the signal-to-noise ratio (SNR) is 20 dB and repeat the first set of experiments.
This experiment was meant to assess the robustness of R-SBL variants under noisy conditions.

In the fourth set of experiments, we investigate recovery performances for a variety of distributions for $\vect{\Phi}$, and for the non-zero elements of $\vect{x}$. 
We randomly draw the nonzero elements of $\vect{x^{gen}}$ according to the following distributions:
\begin{enumerate}[I.]
	\item NN-Cauchy (Location: 0, Scale: 1)
	\item NN-Laplace (Location: 0, Scale: 1)
	\item Gamma (Location: 1, Scale: 2)
	\item Chi-square with $\nu=2$
	\item Bernoulli with $p(0.25) = 1/2$ and $p(1.25)=1/2$	
\end{enumerate}
where the prefix `NN' stands for non-negative. 
These distributions are obtained by taking the absolute value of the respective probability densities. 
We also generate random dictionaries $\vect{\Phi}$ according to the following densities:
\begin{enumerate}[I.]
	\item Normal (Location: 0, Scale: 1)
	\item $\pm 1$ with $p(1) = 1/2$ and $p(-1)=1/2$
	\item $\{0,1\}$ with $p(0) = 1/2$ and $p(1)=1/2$
\end{enumerate}

In all of the experiments detailed here, the results were averaged over 1,000 simulations. 
Moreover, the R-SBL MCMC approach was only used in the first set of experiments to demonstrate the high quality of the parameter estimates obtained with the lower complexity approaches such as DA, LMMSE and GAMP. 
We omit the MCMC in other experiments due to computational constraints.

\vspace{-0.01em}
\subsection{Performance metrics}
To evaluate the performance of various S-NNLS algorithms, we used the normalized mean square error (NMSE) and the probability of error in the recovered support set (PE) \cite{eladbook}.
We computed the NMSE between the recovered signal $\hat{\vectx}$ and the ground truth $\vectx^{gen}$ using
\begin{align}
\text{NMSE} = \|\vect{\hat{x}} -\vect{x^{gen}}\|^2/ \|\vect{x^{gen}}\|^2.
\end{align}
The PE metric was computed using
\begin{align}
\text{PE} = \dfrac{max\{|S|,|\hat{S}|\}- |S\cap \hat{S}|}{max\{|S|,|\hat{S}|\}},
\end{align}
where the support of the true solution was $S$ and the recovered support of $\hat{\vectx}$ was $\hat{S}$.
A value of PE $=0$ indicates that the ground truth and recovered supports are the same, 
whereas PE $=1$ indicates no overlap between supports.
Averaging the PE over multiple trials gives the empirical probability of making errors in the recovered support.  
The averaged values of NMSE and PE over 1,000 simulations and for each experiment are reported in the Experiment Results section.
\begin{figure*}[h!]
\centering
\includegraphics[scale =.3905]{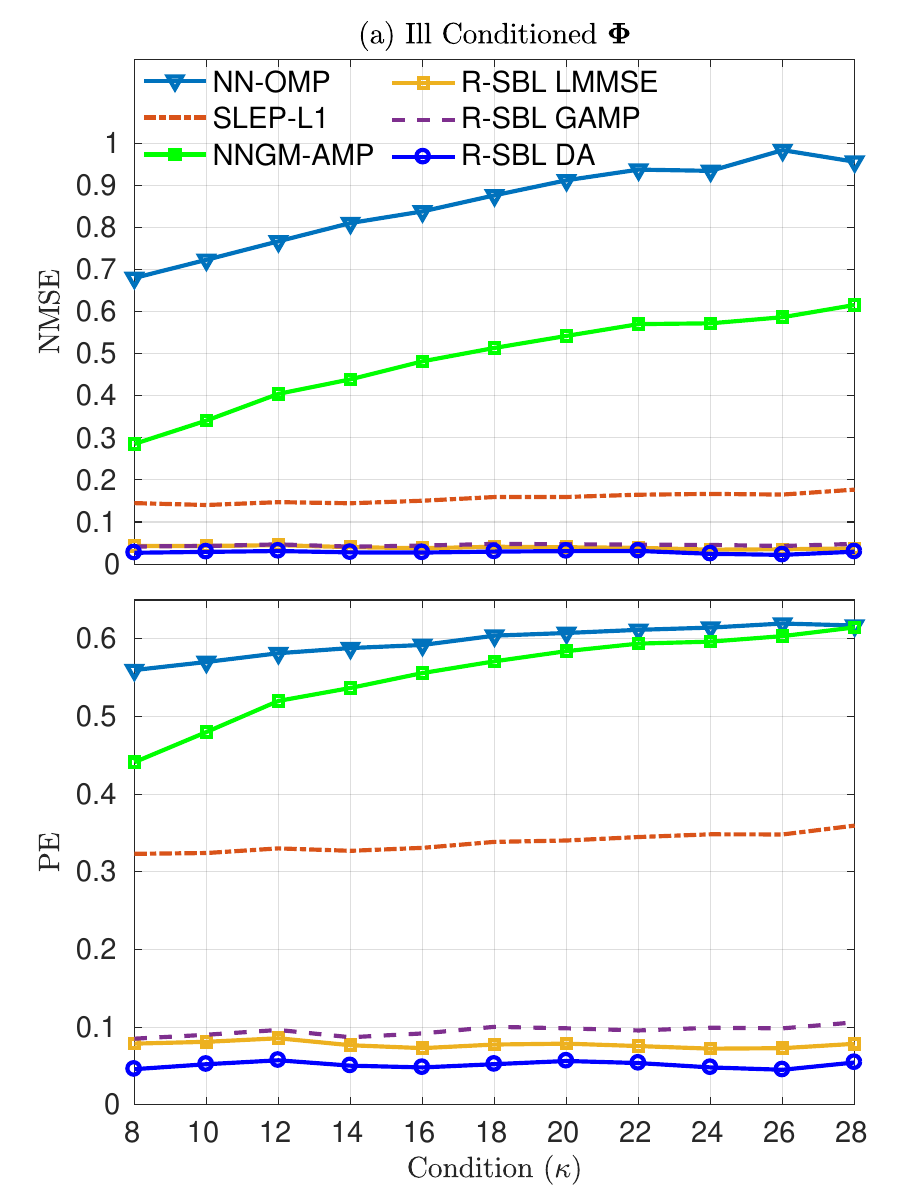}
\includegraphics[scale =.3905]{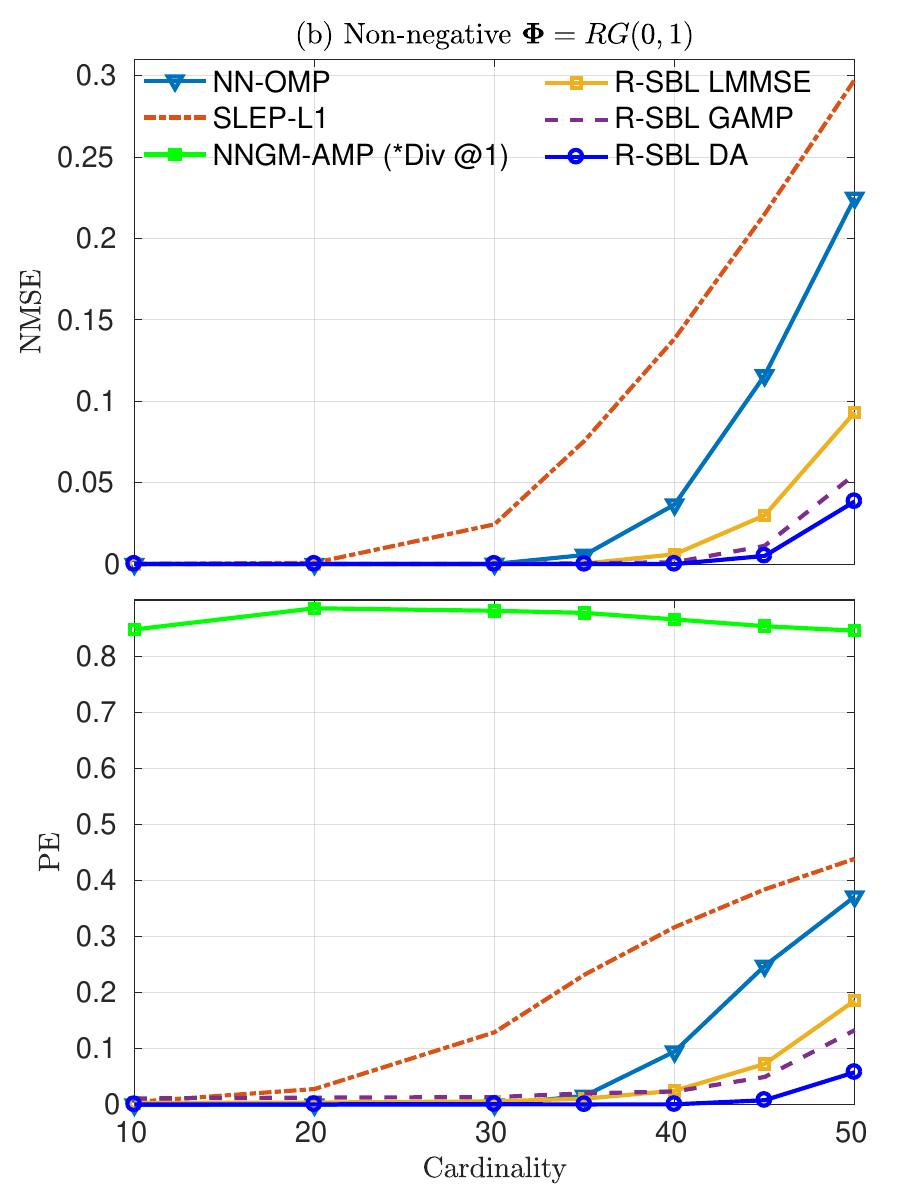}
\includegraphics[scale =.3905]{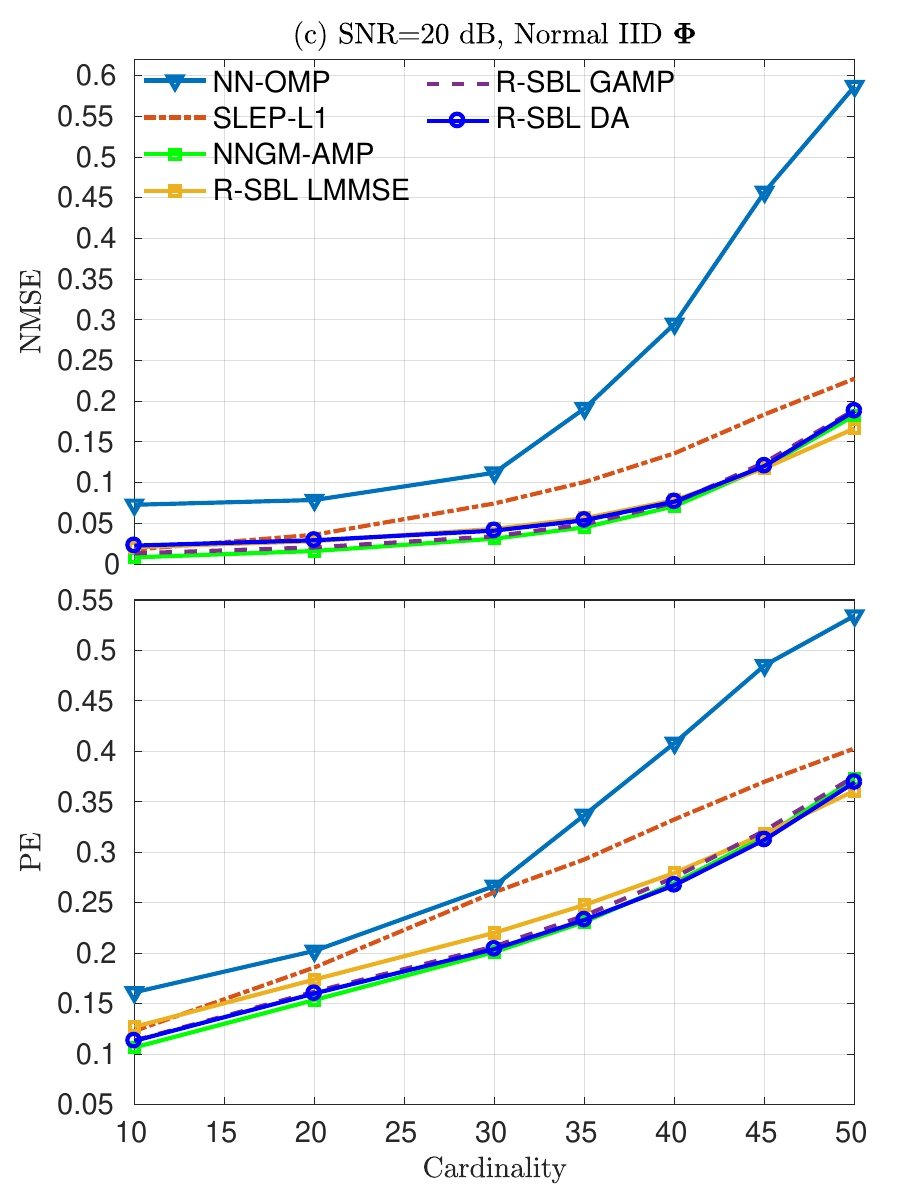}
\caption{
Sparse recovery performances of the S-NNLS solvers for various $\vect{\Phi}$.
In (a) the dictionary is ill-conditioned with condition number $\kappa$ given in the x-axis.
R-SBL variants outperform the baseline solvers for various $\kappa$ and are very robust to the selection of $\kappa$.
R-SBL DA achieves the lowest NMSE and PE.
SLEP-$\ell_1$ is superior to NNGM-AMP.
In (b) the dictionary is non-negative with elements drawn from i.i.d. RG$(0,1)$. 
The recovery performances are given for various cardinality $K$ in the x-axis. 
R-SBL variants achieve superior recovery across all values of $K$. 
NNGM-AMP diverges regardless of the value of $K$ and is unable to recover a feasible solution.
In (c) the dictionary is i.i.d. Normal and SNR is $20$ dB. 
The R-SBL variants perform similar to NNGM-AMP under noisy conditions, but are superior to SLEP-$\ell_1$ and NN-OMP at larger cardinalities.}
\label{fig_f2_norm}
\end{figure*}

\vspace{-0.01em}
\subsection{MCMC implementation}
We used the MCMC implementation presented in \cite{pakman2014exact}.
The Matlab and R codes are available at \href{url}{https://github.com/aripakman/hmc-tmg}.
The MCMC parameters explained in Section \ref{sec:mcmc} were selected as follows. 
The off-diagonal pruning of the empirical scale parameter $\vect{\hat{\Sigma}}$ was performed with a threshold of $T_p = 5\times 10^{-2}$. 
Diagonal scaling was performed with a factor of $\beta=1.7$, and a shrinkage parameter of $\lambda=0.5$. 
These values were empirically determined to minimize the NMSE for the first set of experiments.

\vspace{-0.01em}
\section{Experiment Results}
\label{sec:results}
Here, we show that in all of the sparse recovery experiments detailed above, the proposed R-SBL variants outperform the baseline solvers in terms of NMSE and PE.
The R-SBL variants outperform the baseline solvers when the dictionary is non-i.i.d., coherent, low-rank, ill-posed or even non-negative, showing the robustness of R-SBL to different characteristics of the dictionary $\vect{\Phi}$.

In Fig. \ref{fig_f1_norm}(a) we show the sparse recovery performance of the R-SBL variants and the baseline solvers as a function of the cardinality for the first set of experiments.
As the cardinality of the ground truth solution increases (after $K=30$) the performances of NN-OMP and SLEP-$\ell_1$ deteriorate both in terms of NMSE and PE. 
On the other hand, R-SBL variants and NNGM-AMP are quite robust with very small recovery error.
For the largest cardinality of $K=50$, we see that R-SBL DA and MCMC outperform other methods.
The DA variant is nearly identical to MCMC in terms of NMSE and PE. 
This is expected since MCMC prunes off-diagonal elements of the scale matrix $\vect{\Sigma}$ iteratively, when they drop below a certain threshold. 

\vspace{-0.01em}
\subsection{Coherent Dictionaries}
In Figure \ref{fig_f1_norm}(b) we show the recovery performances when the dictionary is coherent. 
The degree of dictionary coherence is shown on the horizontal axis with $\rho$ which ranges from $0.1$ to $0.95$. 
The proposed R-SBL variants are extremely robust to increasing coherence and outperform the baseline solvers in terms of both NMSE and PE. 
SLEP-$\ell_1$ is robust to increasing coherence but performs worse when compared to the R-SBL variants. 
NNGM-AMP breaks down after $\rho=0.3$ and performs worse than SLEP-$\ell_1$ after $\rho=0.5$, and worse than NN-OMP after $\rho=0.8$.
The LMMSE and DA variants are not affected by the coherence level and achieve better recovery even for $\rho=0.95$.
The performance of R-SBL GAMP slightly deteriorates after an extreme coherence of $\rho=0.90$, but is still better than the baseline solvers.

These results demonstrate that the proposed R-SBL variants are robust to dictionary coherence and are superior to the baseline solvers.
The robustness of our R-SBL framework seems to be inherited from the robustness of the original SBL algorithm to the structure of $\vect{\Phi}$ \cite{al2018gamp,li2013robust}, which uses a GSM prior on $\vect{x}$. Our R-GSM prior on $\vect{x}$ seems to provide a similar robustness to the R-SBL algorithm.

\vspace{-0.01em}
\subsection{Low-rank Dictionaries}
In Figure \ref{fig_f1_norm}(c) we show the recovery performances for rank-deficient dictionaries.
The degree of rank deficiency is shown on the horizontal axis with the rank ratio $R/N$. 
The R-SBL variants outperform the baseline solvers in terms of both NMSE and PE for all values of $R/N$.
The recovery performances of the R-SBL variants are extremely robust against the changes in $R/N$.
Among the R-SBL variants, DA performs slightly better than LMMSE and GAMP, and GAMP performs similar to LMMSE. 
The recovery performance of NNGM-AMP is better than NN-OMP and SLEP-$\ell_1$, however its performance degrades as $R/N$ gets smaller.

\begin{table*}[t!]
    \begin{minipage}{.49\linewidth}
      \centering
\resizebox{1\columnwidth}{!}{%
\setlength\tabcolsep{3pt}
\begin{tabular}{cc|cccccc|}
\cline{3-8}
\multicolumn{1}{l}{} & \multicolumn{1}{l|}{} & \multicolumn{6}{c|}{$\vect{\Phi}$ is i.i.d Normal} \\ \cline{2-8} 
\multicolumn{1}{c|}{} & $\vectx^{gen}$ & NN-OMP & SLEP-$\ell_1$ & \begin{tabular}[c]{@{}c@{}}NNGM \\ AMP\end{tabular} & \begin{tabular}[c]{@{}c@{}}R-SBL \\ (LMMSE)\end{tabular} & \begin{tabular}[c]{@{}c@{}}R-SBL \\ (GAMP)\end{tabular} & \begin{tabular}[c]{@{}c@{}}R-SBL \\ (DA)\end{tabular} \\ \hline
\multicolumn{1}{|c|}{\multirow{6}{*}{\rot{NMSE}}} & RG & 0.4460 & 0.1439 & 0.0389 & 0.0488 & 0.0428 & \textbf{0.0313} \\
\multicolumn{1}{|c|}{} & NN-Cauchy & 0.0097 & 0.0086 & 0.0020 & 0.0004 & 0.0003 & \textbf{0.0002} \\
\multicolumn{1}{|c|}{} & NN-Laplace & 0.1566 & 0.0693 & 0.0091 & 0.0066 & 0.0059 & \textbf{0.0034} \\
\multicolumn{1}{|c|}{} & Gamma & 0.1476 & 0.0661 & 0.0074 & 0.0065 & 0.0045 & \textbf{0.0024} \\
\multicolumn{1}{|c|}{} & Chi-square & 0.1583 & 0.0673 & 0.0091 & 0.0077 & 0.0066 & \textbf{0.0035} \\
\multicolumn{1}{|c|}{} & Bernoulli & 0.5845 & 0.1265 & \textbf{0.0052} & 0.0524 & 0.0416 & 0.0339 \\ \hline
\multicolumn{1}{|c|}{\multirow{6}{*}{\rot{PE}}} & RG & 0.4601 & 0.3208 & 0.0711 & 0.0873 & 0.0823 & \textbf{0.0549} \\
\multicolumn{1}{|c|}{} & NN-Cauchy & 0.2307 & 0.3509 & 0.2142 & \textbf{0.0187} & 0.0200 & 0.0408 \\
\multicolumn{1}{|c|}{} & NN-Laplace & 0.3202 & 0.3137 & 0.0407 & 0.0292 & 0.0229 & \textbf{0.0118} \\
\multicolumn{1}{|c|}{} & Gamma & 0.3091 & 0.3093 & 0.0416 & 0.0260 & 0.0207 & \textbf{0.0080} \\
\multicolumn{1}{|c|}{} & Chi-square& 0.3200 & 0.3086 & 0.0473 & 0.0307 & 0.0280 & \textbf{0.0133} \\
\multicolumn{1}{|c|}{} & Bernoulli & 0.4852 & 0.3283 & \textbf{0.0101} & 0.1714 & 0.1514 & 0.1264 \\ \hline
\end{tabular}}
      \caption{NMSE and PE results for various distributions for $\vectx^{gen}$. The dictionary is i.i.d Normal distributed.}\label{table1}
    \end{minipage}%
    \hfill
    \begin{minipage}{.49\linewidth}
      \centering
\resizebox{1\columnwidth}{!}{%
\setlength\tabcolsep{3pt}
\begin{tabular}{cc|cccccc|}
\cline{3-8}
\multicolumn{1}{l}{} & \multicolumn{1}{l|}{} & \multicolumn{6}{c|}{$\vect{\Phi}$ is $\pm 1$ Bernoulli} \\ \cline{2-8} 
\multicolumn{1}{c|}{} & $\vectx^{gen}$ & NN-OMP & SLEP-$\ell_1$ & \begin{tabular}[c]{@{}c@{}}NNGM \\ AMP\end{tabular} & \begin{tabular}[c]{@{}c@{}}R-SBL \\ (LMMSE)\end{tabular} & \begin{tabular}[c]{@{}c@{}}R-SBL \\ (GAMP)\end{tabular} & \begin{tabular}[c]{@{}c@{}}R-SBL \\ (DA)\end{tabular} \\ \hline
\multicolumn{1}{|c|}{\multirow{6}{*}{\rot{NMSE}}} & RG & 0.3996 & 0.1387 & 0.0409 & 0.0504 & 0.0415 & \textbf{0.0332} \\
\multicolumn{1}{|c|}{} & NN-Cauchy & 0.0083 & 0.0077 & 0.0023 & 0.0005 & 0.0004 & \textbf{0.0003} \\
\multicolumn{1}{|c|}{} & NN-Laplace & 0.1368 & 0.0712 & 0.0101 & 0.0096 & 0.0090 & \textbf{0.0050} \\
\multicolumn{1}{|c|}{} & Gamma & 0.1294 & 0.0665 & 0.0079 & 0.0061 & 0.0051 & \textbf{0.0023} \\
\multicolumn{1}{|c|}{} & Chi-square & 0.1267 & 0.0667 & 0.0109 & 0.0083 & 0.0094 & \textbf{0.0055} \\
\multicolumn{1}{|c|}{} & Bernoulli & 0.5610 & 0.1180 & \textbf{0.0113} & 0.0466 & 0.0412 & 0.0363 \\ \hline
\multicolumn{1}{|c|}{\multirow{6}{*}{\rot{PE}}} & RG & 0.4272 & 0.3182 & 0.0794 & 0.0950 & 0.0824 & \textbf{0.0568} \\
\multicolumn{1}{|c|}{} & NN-Cauchy & 0.1810 & 0.3475 & 0.2307 & 0.0187 & \textbf{0.0175} & 0.0321 \\
\multicolumn{1}{|c|}{} & NN-Laplace & 0.2909 & 0.3131 & 0.0508 & 0.0369 & 0.0333 & \textbf{0.0163} \\
\multicolumn{1}{|c|}{} & Gamma & 0.2682 & 0.3072 & 0.0472 & 0.0274 & 0.0248 & \textbf{0.0093} \\
\multicolumn{1}{|c|}{} & Chi-square & 0.2769 & 0.3104 & 0.0532 & 0.0357 & 0.0369 & \textbf{0.0195} \\
\multicolumn{1}{|c|}{} & Bernoulli & 0.4734 & 0.3290 & \textbf{0.0154} & 0.1727 & 0.1571 & 0.1345 \\ \hline
\end{tabular}}
 \caption{NMSE and PE results for various distributions for $\vectx^{gen}$. The dictionary is i.i.d $\pm 1$ Bernoulli distributed.}\label{table2}
    \end{minipage} 
\end{table*}

\vspace{-0.01em}
\subsection{Ill-conditioned Dictionaries}
In Figure \ref{fig_f2_norm}(a) we demonstrate the recovery performances for ill-conditioned dictionaries. 
The condition number on the horizontal axis varies from $\kappa=8$ to $\kappa=28$.
The proposed R-SBL variants perform significantly better than the baseline solvers across different $\kappa$ values in terms of NMSE and PE.
The recovery performances of the R-SBL variants are also extremely robust to different selections of $\kappa$.
SLEP-$\ell_1$ is better than NN-OMP and NNGM-AMP and is also robust to the selection of $\kappa$.
The performances of NN-OMP and NNGM-AMP methods rapidly deteriorate with increasing $\kappa$ values.

\vspace{-0.01em}
\subsection{Non-negative Dictionaries}
In Figure \ref{fig_f2_norm}(b) we show the recovery performances when the dictionary is non-negative with elements drawn from i.i.d. $RG(0,1)$. 
The cardinality $K$ on the horizontal axis of Figure \ref{fig_f2_norm}(b) varies from $K=10$ to $K=50$.
The NNGM-AMP approach was not able to recover feasible solutions for non-negative dictionaries and the point estimates for $\vect{x}$ diverged for different $K$. 
Therefore, the NMSE values for NNGM-AMP were not shown in Figure \ref{fig_f2_norm}(b).
Unlike in Figure \ref{fig_f1_norm}(a), where the dictionary can be both positive and negative, NN-OMP performs better than SLEP-$\ell_1$. 
The proposed R-SBL variants outperform the baseline approaches.
Among the R-SBL variants, DA performs slightly better than GAMP, and GAMP is slightly better than LMMSE.

\vspace{-0.01em}
\subsection{Noisy Conditions}
We compared the recovery performances in a noisy setting, where the dictionary is i.i.d. Normal distributed.
In this case, the observations were contaminated with additive white Gaussian noise to have a signal-to-noise ratio (SNR) of 20 dB.
Figure \ref{fig_f2_norm}(c) shows the NMSE and PE versus the cardinality.
Compared with the noiseless case in Figure \ref{fig_f1_norm}(a), the performances of all of the methods noticeably reduced.
However, the proposed R-SBL variants performed better as compared to the NN-OMP and SLEP-$\ell_1$ solvers, and performed similar to the NNGM-AMP approach.

\vspace{-0.01em}
\subsection{Other types of $\vectx^{gen}$ and $\vect{\Phi}$}
Here, the dictionary $\vect{\Phi}$ was drawn according to i.i.d. Normal, $\pm 1$ Bernoulli, and $\{0,1\}$ Bernoulli distributions. 
We experimented with different distributions for the non-zero entries of $\vectx^{gen}$, as detailed in Tables \ref{table1}, \ref{table2} and \ref{table3}.

For i.i.d. Normal $\vect{\Phi}$ in Table \ref{table1}, the R-SBL DA generally outperforms the baseline solvers and other R-SBL variants when $\vectx^{gen}$ is RG, NN-Cauchy, NN-Laplace, Gamma and Chi-square distributed. 
The LMMSE variant achieves slightly better performance in terms of PE for the NN-Cauchy distribution. 
The NNGM-AMP is better than LMMSE and GAMP variants, when $\vectx^{gen}$ is RG, however it fails in terms of PE when $\vectx^{gen}$ is NN-Cauchy. 
The NNGM-AMP approach shows better performance when $\vectx^{gen}$ is Bernoulli. 
This is expected since the prior density for NNGM-AMP is a Bernoulli non-negative Gaussian mixture. 
The R-GSM prior, on the other hand, is not well matched to the Bernoulli distribution, as it is a mixture of continuous distributions. 
Overall, we see that R-SBL DA approach results in the best recovery performance. 

In Table \ref{table2}, we present the results for when $\vect{\Phi}$ is $\pm 1$ Bernoulli. 
The recovery performances observed in Table \ref{table2} are very similar to Table \ref{table1} and overall, the R-SBL DA approach enjoys better recovery performance.

In Table \ref{table3}, we show recovery results for $\{0,1\}$ Bernoulli distributed $\vect{\Phi}$.
The R-SBL DA and LMMSE variants achieve superior recovery when compared to the baseline solvers. 
The NNGM-AMP approach diverges for different $\vectx^{gen}$.
This is consistent with our previous observation that NNGM-AMP failed when the dictionary elements were positive e.g. drawn from i.i.d. $RG(0,1)$ in Figure \ref{fig_f2_norm}(b).

\begin{table}[!b]
      \centering
\resizebox{1\columnwidth}{!}{%
\setlength\tabcolsep{3pt}
\begin{tabular}{cc|cccccc|}
\cline{3-8}
\multicolumn{1}{l}{} & \multicolumn{1}{l|}{} & \multicolumn{6}{c|}{$\vect{\Phi}$ is $\{0,1\}$ Bernoulli} \\ \cline{2-8} 
\multicolumn{1}{c|}{} & $\vectx^{gen}$ & NN-OMP & SLEP-$\ell_1$ & \begin{tabular}[c]{@{}c@{}}NNGM \\ AMP\end{tabular} & \begin{tabular}[c]{@{}c@{}}R-SBL \\ (LMMSE)\end{tabular} & \begin{tabular}[c]{@{}c@{}}R-SBL \\ (GAMP)\end{tabular} & \begin{tabular}[c]{@{}c@{}}R-SBL \\ (DA)\end{tabular} \\ \hline
\multicolumn{1}{|c|}{\multirow{6}{*}{\rot{NMSE}}} & RG & 0.2063 & 0.2497 & Diverged & 0.0873 & 0.0520 & \textbf{0.0386} \\
\multicolumn{1}{|c|}{} & NN-Cauchy & 0.0085 & 0.0188 & Diverged & 0.0031 & 0.0286 & \textbf{0.0002} \\
\multicolumn{1}{|c|}{} & NN-Laplace & 0.0960 & 0.1406 & Diverged & 0.0296 & 0.0070 & \textbf{0.0043} \\
\multicolumn{1}{|c|}{} & Gamma & 0.0901 & 0.1335 & Diverged & 0.0283 & 0.0047 & \textbf{0.0022} \\
\multicolumn{1}{|c|}{} & Chi-square & 0.0894 & 0.1360 & Diverged & 0.0327 & 0.0077 & \textbf{0.0054} \\
\multicolumn{1}{|c|}{} & Bernoulli & 0.2203 & 0.2586 & Diverged & 0.0747 & 0.0682 & \textbf{0.0558} \\ \hline
\multicolumn{1}{|c|}{\multirow{6}{*}{\rot{PE}}} & RG & 0.3558 & 0.4070 & 0.8404 & 0.1782 & 0.0950 & \textbf{0.0581} \\
\multicolumn{1}{|c|}{} & NN-Cauchy & 0.2651 & 0.4434 & 0.8314 & 0.1131 & 0.4480 & \textbf{0.0354} \\
\multicolumn{1}{|c|}{} & NN-Laplace & 0.3140 & 0.4071 & 0.8398 & 0.1193 & 0.0371 & \textbf{0.0134} \\
\multicolumn{1}{|c|}{} & Gamma & 0.3102 & 0.4016 & 0.8354 & 0.1240 & 0.0275 & \textbf{0.0087} \\
\multicolumn{1}{|c|}{} & Chi-square & 0.3126 & 0.4072 & 0.8377 & 0.1341 & 0.0363 & \textbf{0.0171} \\
\multicolumn{1}{|c|}{} & Bernoulli & 0.3803 & 0.4120 & 0.8399 & 0.2689 & 0.1705 & \textbf{0.1455} \\ \hline
\end{tabular}}
 \caption{NMSE and PE results for various distributions for $\vectx^{gen}$. The dictionary is i.i.d  $\{0,1\}$ Bernoulli distributed.}\label{table3}
\end{table}

\vspace{-0.01em}
\subsection{Recovery time analysis}
In Section \ref{time_comp}, we presented the worst case computational complexity of the DA, LMMSE and GAMP variants per EM iteration.
As the execution time also depends on how fast an EM approach converges to the final solution, we provide an analysis of the average execution times for different cardinality values.
First, we provide a simple way to speed up the proposed R-SBL algorithms.
We prune the problem size when the elements of $\vect{\gamma}$ become smaller than a given threshold.
For example, when an index of the vector $\vect{\gamma}$ becomes smaller than i.e. ${\gamma}_i \leq \epsilon_{\vect{\gamma}}$, we ignore the computations regarding that index in the next iterations. 
This effectively reduces the problem dimensions and improves execution time.

\begin{figure}[t!]
\centering
\includegraphics[width=0.4\textwidth]{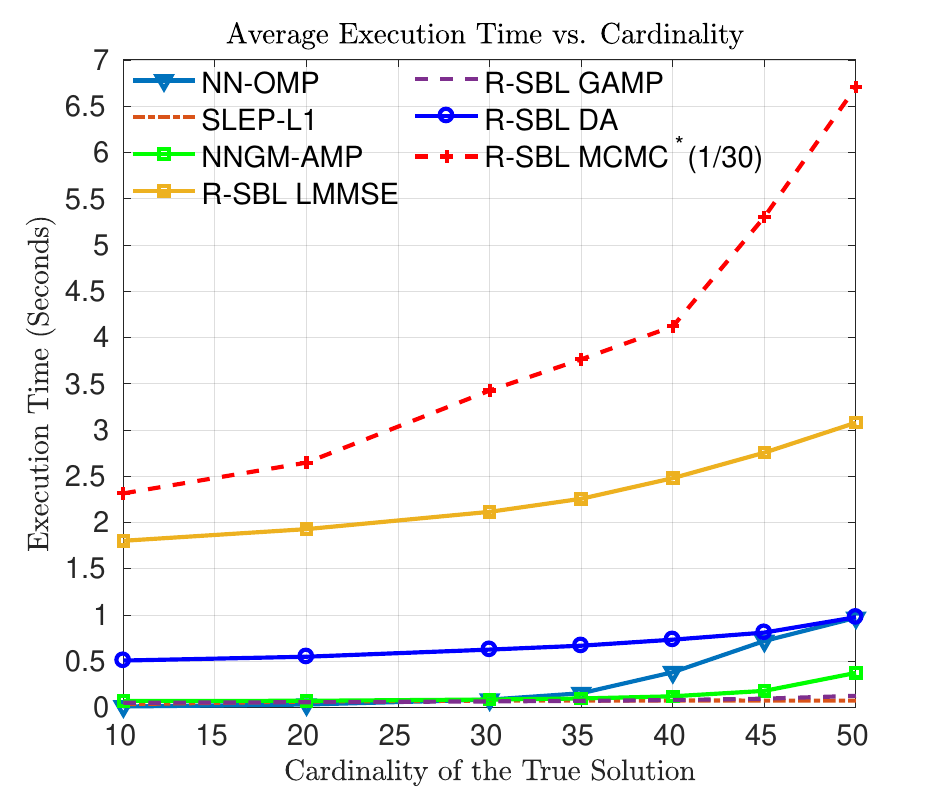}
\caption{Execution times of the S-NNLS solvers as a function of cardinality for the noiseless scenario.}
\label{fig_f4_timing}
\end{figure}

In Fig. \ref{fig_f4_timing}, we included the average execution times of the proposed algorithms in units of seconds. 
The pruning threshold was selected as $\epsilon_{\vect{\gamma}} = 10^{-5}$ for all methods.
For the EM based methods, we monitored the convergence of the $\vect{\gamma}$'s in EM iterations. 
We stopped the EM updates when $\|\vect{\gamma}^{t} -\vect{\gamma}^{t-1}\|_2\leq 10^{-3}$, where $t$ is the current EM iteration index. 
For other approaches, we monitored the linear equality constraints and stopped the algorithms when $\|\vect{y}-\vect{\Phi}\vect{\hat{x}^{t}} \|_2 \leq 10^{-3}$, where $\vect{\hat{x}}^{t}$ is the solution estimate at iteration $t$.

As expected due to computationally intensive random sampling, R-SBL MCMC is the slowest method. 
For display purposes, we scaled down the average MCMC execution time values by 30.
The LMMSE approach takes about 3 seconds for $K=50$ to recover the optimal solution and is the second slowest method.
Even though the complexity of DA and LMMSE is similar, DA achieves much faster convergence and takes about 0.5 to 1 seconds as $K$ increases. 

For this particular experiment, GAMP is the fastest R-SBL variant regardless of the cardinality and is similar to SLEP-$\ell_1$. 
However, since the complexity of GAMP is $\mathcal{O}(NM)$, for very large problem sizes (e.g. large $N$ and $M$) GAMP may become slower despite superior recovery performance. 
In this case, a convex solver may be preferable depending on the desired recovery performance.
R-SBL GAMP is faster than NNGM-AMP at larger cardinalities.
Finally, NN-OMP is similar to SLEP-$\ell_1$ but its execution time increases for larger cardinalities.
Considering the fast recovery speed and good recovery performance of R-SBL GAMP under various $\vect{\Phi}$ types, the R-SBL GAMP variant is a very good candidate for time sensitive sparse recovery applications. 

\vspace{-0.01em}
\subsection{Application on real data: Face Recognition}
Here, we present a face recognition (FR) application based on the non-negative sparse representations considered in \cite{he2013two,he2011maximum,vo2009nonnegative}.  
Our goal is to show that the R-SBL approach works well in real-world applications involving real-data.
A sparse representation classifier (SRC) for FR was initially proposed in \cite{wright2009robust} using the $\ell_1$ penalty without the non-negativity constraints. 
The SRC approach was found to be robust against occlusion, disguise, pixel corruptions, and achieved superior results as compared to well-known FR algorithms \cite{wright2009robust,he2011maximum,turk1991eigenfaces,he2005face}. 

In the SRC framework, the dictionary $\vect{\Phi}$ represents the training samples and each column of $\vect{\Phi}$ contains training features from a single face image. 
A single person may have more than one training image, and hence multiple columns of $\vect{\Phi}$ might correspond to the same person.
For a given test face $\vect{y}$ in vectorized form, a vector $\vect{x}$ is obtained by solving Eq. \eqref{intro_1} using $\ell_1$ sparsity, with the assumption that only a few non-zero entries will exist in the solution $\vect{x}$.
Ideally, the index of the maximal non-negative entry in $\vect{x}$ is used to select the corresponding column in $\vect{\Phi}$. 
This column should correspond to one of the training samples for the correct person.
In \cite{he2011maximum}, the SRC performance was further improved by adding the non-negativity constraint on $\vect{x}$ in addition to the $\ell_1$ sparsity. 
The authors have shown their algorithm to be more robust against noise and to be computationally more efficient as compared to the original SRC approach.
\begin{figure*}[t!]
\centering
\includegraphics[width=0.49\textwidth]{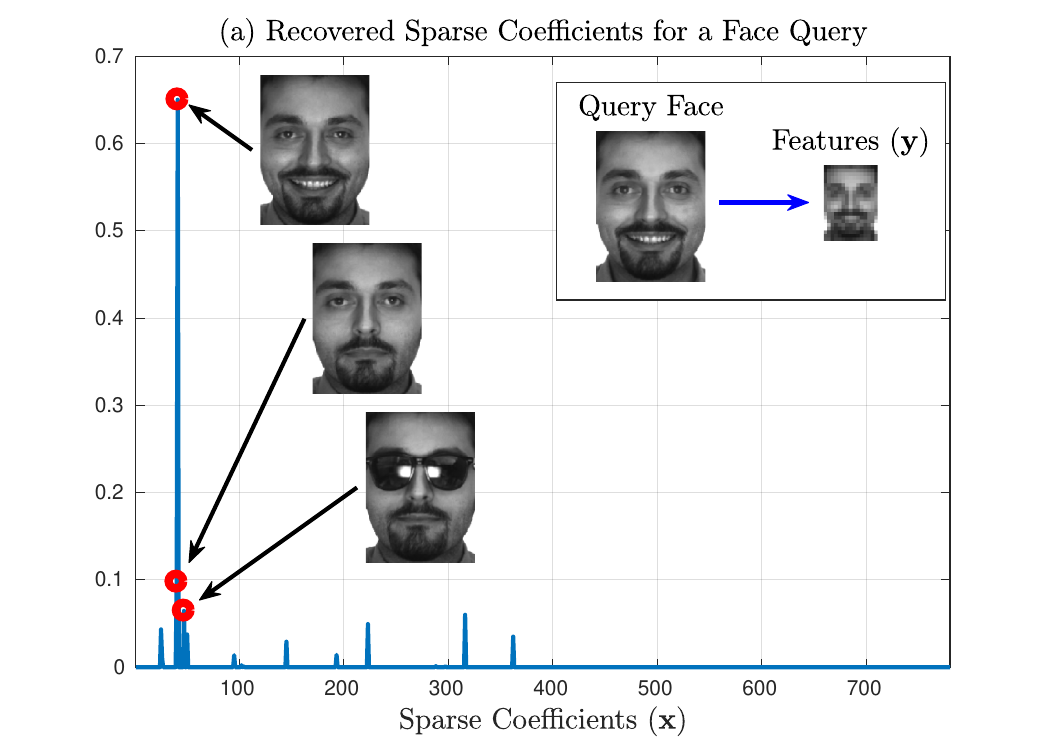}
\includegraphics[width=0.49\textwidth]{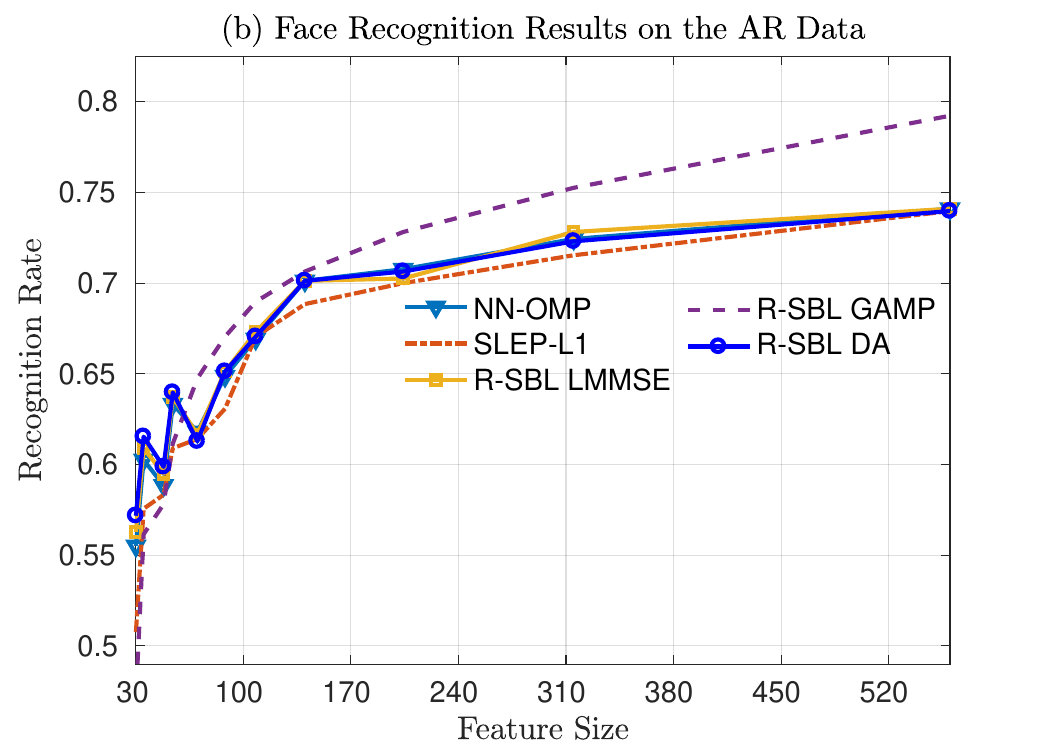}
\caption{(a) Illustration of the sparse FR process. A query face is down-sampled to obtain an observation $\vect{y}$. Using the training dictionary $\vect{\Phi}$, a sparse solution is obtained using the R-SBL variants and baseline solvers to satisfy $\vect{y}= \vect{\Phi} \vect{x}$. The index that corresponds to the maximum positive value in $\vect{x}$ is used to select a corresponding column in $\vect{\Phi}$. The image in this column corresponds to the correct individual. (b) FR accuracy for different feature sizes using all test samples. The R-SBL GAMP enjoys better FR performance for different feature sizes.}
\label{FR}
\end{figure*}

In our experiment, we consider the R-SBL framework for the FR problem and compare it with the baseline solvers. Note that SLEP-$\ell_1$ was considered as the non-negative $\ell_1$ minimization counterpart of R-SBL in place of \cite{he2011maximum}.
We used the public AR dataset \cite{martinez1998ar} and selected the first 30 males and 30 females for the FR problem. 
Each person in the dataset has 26 face images with different facial expression, illumination, and disguise (e.g. sunglasses and scarves). 
The first 13 images of each person ($M = 13\times 60 = 780$) were selected as the training set, and the remaining $780$ face images were used for testing.
For feature selection, we used the down-sampling method used in \cite{he2013two,he2011maximum,wright2009robust}, where the pixel dimensions of each face image were down-sampled to have a total of $N$ pixels. 
In separate experiments, each $165\times 120$ pixel image was down-sampled by a factor of $\{1/28,1/26,...,1/6\}$, yielding feature dimensions of minimum of  $N=30$ to a maximum of $N=650$. 

The overall process is shown in Figure \ref{FR}(a), where a query face is shown in the top right-hand side panel. 
This image was then down-sampled and the original feature dimension was reduced from $19,800$ to $512$. 
After sparse recovery with R-SBL, the original faces belonging to several largest non-zero elements of $\vect{x}$ are shown. 
As desired, the maximal positive index of $\vect{x}$ belongs to the same person in the query face.

In Figure \ref{FR}(b) we performed FR using all $780$ samples in the test set and measured the recognition rate for different feature sizes. 
The recognition rate was computed by counting the number of test samples for which R-SBL recovered the correct individual from $\vect{x}$. 
This count was normalized by $780$.
Overall, the R-SBL variants with the exception of R-SBL GAMP performed similar to the baseline solvers for large feature sizes. 
This is expected since the recovery problem was highly sparse, and the cardinality was very small $K=13$ as compared to the length of $\vect{x}$ (i.e. largest length of  $\vect{x}$ is 780). 
R-SBL GAMP was superior to all algorithms for large feature sizes and performed significantly better in identifying the correct individual. 
NNGM-AMP diverged for this application and did not yield reportable results.

\section{Conclusion}
\label{sec:conclusion}
In this work we introduced a hierarchical Bayesian method to solve the S-NNLS problem.
We proposed the rectified Gaussian scale mixture model as a general and versatile prior to promote sparsity in the solution of interest.
Since the marginals of the posterior were not tractable, we constructed our R-SBL algorithm using the EM framework with four different approaches.
We demonstrated that our R-SBL approaches outperformed the available S-NNLS solvers, in most cases by a large margin. 
The proposed R-SBL framework is very robust to the structure of $\vect\Phi$ and performed well regardless of $\vect\Phi$ being i.i.d. and non-i.i.d. distributed.
The performance gains achieved by the R-SBL variants are consistent across different non-negative data distributions for $\vectx$,  and different structures for the design matrix $\vect\Phi$ in coherent, low-rank, ill-posed and non-negative settings. 
The DA variant was found to be an easy to implement S-NNLS solver with simple closed-form moment expressions.

\section{Appendix}
\subsection{Full derivation of GAMP}\label{sec:gamp}
We use the R-GSM prior $p(\boldsymbol{x} \mathcal{\vert} \boldsymbol{\gamma})$ and evaluate Eq. \eqref{x_sum_product} and Eq. \eqref{var_sum_product} to find the first two moments of the approximate marginal posterior under the sum-product GAMP mode
\begin{flalign}
\hat{x}_i&=\mathbb{E}\{ x_i \mathcal{\vert} r_i;\tau_{r_i} \} = \int_{x_i} x_i p(x_i \mathcal{\vert} r_i;\tau_{r_i})\\
&= \int_+ x_i  \mathcal{N}^R (x_i \mathbb{\vert} 0,\gamma_i) \mathcal{N} (x_i,r_i,\tau_{r_i}),
\end{flalign}
then using the Gaussian multiplication rule\footnotemark{,} we obtain
\footnotetext{$\mathcal{N}(x;\mu_a,\tau_a) \mathcal{N}(x;\mu_b,\tau_b)=\Upsilon \mathcal{N}(x;\frac{\frac{\mu_a}{\tau_a}+\frac{\mu_b}{\tau_b}}{\frac{1}{\tau_a}+\frac{1}{\tau_b}},\frac{1}{\frac{1}{\tau_a}+\frac{1}{\tau_b}}),$ where $\Upsilon$ is a scaling factor.}
\begin{flalign}
\hat{x}_i&= \int_+ x_i \Upsilon \mathcal{N}^R (x_i \mathbb{\vert} \eta_i,\nu_i),
\end{flalign}
where $\eta_i$ and $\nu_i$ are given in Eq. \eqref{eta} and Eq. \eqref{nu}, respectively. 

We then find the mean of the resulting rectified Gaussian
\begin{flalign}
&\hat{x}_i= \eta_i+\sqrt{\nu_i} h(\frac{\eta_i}{\nu_i}) \label{mean}\\
&\eta_i=\frac{r_i \gamma_i}{\tau_{r_i}+\gamma_i} \label{eta}\\
&\nu_i=\frac{\tau_{r_i} \gamma_i}{\tau_{r_i}+\gamma_i} \label{nu}\\
&h(a)=\frac{\varphi(a)}{\Phi_c(a)},
\end{flalign}
where $\varphi$ refers to the pdf and $\Phi_c$ refers to the complementary cdf of a zero-mean and unit-variance Gaussian distribution. The conditional variance of $x_i$ given $r_i$ is simply
\begin{flalign}
\tau_{x_i}&=var\{ x_i \mathcal{\vert} r_i;\tau_{r_i} \} = \int_{x_i} x^2_i p(x_i \mathcal{\vert} r_i;\tau_{r_i}) - \hat{x}^2_i\\
&= \int_+ x^2_i \mathcal{N}^R (x_i \mathbb{\vert} 0,\gamma_i) \mathcal{N} (x_i,r_i,\tau_{r_i})- \hat{x}^2_i,
\end{flalign}
using the Gaussian multiplication rule
\begin{flalign}
\tau_{x_i}= \int_+ x^2_i \Upsilon \mathcal{N}^R (x_i \mathbb{\vert} \eta_i,\nu_i),
\end{flalign}
we find the variance of the resulting rectified Gaussian as
\begin{flalign}
\tau_{x_i}&= \nu_i g(\frac{\eta_i}{\nu_i}) \label{variance}\\
g(a)&=1-h(a) \left(h(a)-a \right).
\end{flalign}
In the case of max-sum GAMP implementation, we evaluate Eq. \eqref{x_max_sum} and Eq. \eqref{var_max_sum}
\begin{flalign}
\hat{x}_i &=\argmin_{\hat{x}_i \geq 0} \frac{x_i^2}{2 \gamma_i} + \frac{1}{2 \tau_{r_i}} |\hat{x}_i-r_i|^2\\
\hat{x}_i&=
\begin{cases}
\frac{r_i \gamma_i}{\tau_{r_i}+\gamma_i}=\eta_i  &\text{if  } \hat{x}_i \geq 0 \\
0 &\text{if  } \hat{x}_i < 0
\end{cases}
\end{flalign}
Using Eq. \eqref{var_max_sum}
\begin{flalign}
\tau_{x_i}&=
\begin{cases}
\frac{\tau_{r_i} \gamma_i}{\tau_{r_i}+\gamma_i}=\nu_i &\text{if  } \hat{x}_i \geq 0 \\
0 &\text{if  } \hat{x}_i < 0 \footnotemark
\end{cases}
\end{flalign}
\footnotetext{Practically it was found that setting $\tau_{x_i} = 0$ when $\hat{x}_i < 0$ increases the chances of the algorithm getting stuck at a local minimum. Instead, we set $\tau_{x_i}=\frac{\tau_{r_i} \gamma_i}{\tau_{r_i}+\gamma_i}=\nu_i$.}
Upon convergence of the max-sum, the approximate marginals are obtained using Eq. \eqref{mean} and Eq. \eqref{variance}.

\subsection{Approximate marginals and moments using DA}\label{sec:posterior}
We derive the approximate moments used in the R-SBL DA approximation.
We start with the posterior $p(\vectx | \vecty,\vectg)$ and use chain rule to write
\begin{equation}
\label{eq:bayes}
\begin{split}
p(\vectx | \vecty,\vectg) =\dfrac{p(\vecty|\vectx,\vectg)p(\vectx|\vectg)}{\int_{\vectx} p(\vecty|\vectx,\vectg)p(\vectx|\vect{\gamma})d \vectx}.
\end{split}
\end{equation}
Here $p(\vecty | \vectx,\vectg)$ is a Gaussian density due to the Gaussian noise assumption. 
Since $p(\vectx|\vectg)$ is a rectified Gaussian density the numerator of Eq. \eqref{eq:bayes} is a Gaussian multiplied by a rectified Gaussian, which results in a rectified Gaussian density. 
Then, we can simply write
\begin{align}
p(\vectx|\vecty,\vectg) = c(\vect{y}) e^{-\dfrac{(\vectx-\vect{\mu})^T \vect\Sigma^{-1}(\vectx-\vect{\mu})}{2}}u(\vectx),
\end{align}
where $c(\vect{y})$ is the normalizing constant for the posterior density and $\vect{\mu}$ and $\vect\Sigma$ are given by Eqs. \eqref{mu} and \eqref{sigma}, respectively.
Let $\vect\Sigma=\vect{L}\vect{L}^T$ and $\vect{r}=\vectx-\vect{\mu}$, so that $d\vectx =d \vect{r}$ and $\vect\Sigma^{-1}=\vect{L}^{-T}\vect{L}^{-1}$. Therefore, we have
\begin{equation}\label{posterior:p1}
1 = c(\vect{y}) \int_{-\vect{\mu}}^{\infty}e^{\dfrac{-\vect{r}^T\vect{L}^{-T}\vect{L}^{-1}\vect{r}}{2}}d \vect{r}.
\end{equation}
Now, let $\vect{z}=\vect{L}^{-1}\vect{r}$, which implies that $d\vect{r}=|\vect{L}|d\vect{z}$ and
\begin{equation}
\begin{split}
c(\vect{y}) =  \dfrac{1}{ |\vect{L}| \int_{-\vect{\beta}}^{\infty}e^{-\vect{z}^T\vect{z}/2}d\vect{z}},
\end{split}
\label{integral}
\end{equation}
where $\vect{\beta}=\vect{L}^{-1}\vect{\mu}$ is the lower limit of the new integral in vector form.
The lower limit $\vect{\beta}$ depends on a linear combination of elements of $\vect{\mu}$ since $\vect{L}$ is not diagonal.
Thus, the integral in the denominator of Eq. \eqref{integral} is \textbf{not tractable} as the integration limits are not separable and the multidimensional integral over $\vect{z}$ in Eq. \eqref{integral} is not separable as a product of one dimensional integrals.

Assume that, we are interested in an approximate density $\tilde{p}(\vectx | \vecty,\vectg)$, instead of the exact posterior. 
We calculate an approximate $\tilde{c}(\vect{y})$ by approximating $\vect{\Sigma}$ with its diagonal
i.e. $\vect{\Sigma}_d = diag(\vect{\Sigma}) \approx \vect{\Sigma}$. 
In this case, the new $\vect{L}$ is diagonal with entries $\sqrt{{\Sigma}_{ii}}$. 
Thus, the integral in Eq. \eqref{integral} is separable and the approximate normalizing constant $\tilde{c}(\vect{y})$ has closed form
\begin{align}\label{appendix:constant}
\tilde{c}(\vect{y}) = \dfrac{1}{|\vect\Sigma_d|^{1/2} \prod_{i=1}^{M} \sqrt{\dfrac{\pi}{2}} \erfc\left(-\dfrac{\mu_i}{\sqrt{2{{\Sigma}_{ii}}}}\right)}.
\end{align}
Approximating the actual normalizing constant with $\tilde{c}(\vect{y})$, we write the approximate posterior as
\begin{align}
\tilde{p}(\vectx|\vecty,\vectg) =&  \dfrac{e^{-\dfrac{(\vectx-\vect{\mu})^T \vect\Sigma_d^{-1}(\vectx-\vect{\mu})}{2}}u(\vectx)}{
\prod_{i=1}^{M} \sqrt{\dfrac{\pi\Sigma_{ii}}{2}} \erfc\left(-\dfrac{\mu_i}{\sqrt{2{{\Sigma}_{ii}}}}\right)} \\
=& \prod_{i=1}^{M}  \sqrt{\dfrac{2}{\pi\Sigma_{ii}}} \dfrac{e^{-\dfrac{(x_i-\mu_i)^2}{2\Sigma_{ii}}}u(x_i)}{
 \erfc\left(-\dfrac{\mu_i}{\sqrt{2{{\Sigma}_{ii}}}}\right)}\\
 =&  \prod_{i=1}^{M} \tilde{p}(x_i| \vect{y},\vect{\gamma})
\label{appost_3}
\end{align}
Eq. \eqref{appost_3} shows that multivariate $\tilde{p}(\vectx|\vecty,\vectg)$ is separable into product of univariate densities.
The univariate density $\tilde{p}(x_i| \vect{y},\vect{\gamma})$ is the univariate RG density defined in Eq. \eqref{rgg_density} e.g.  $\tilde{p}(x_i| \vect{y},\vect{\gamma}) = \mathcal{N}^{R}(x_i;\mu_i,\Sigma_{ii})$.
The first and second moments of a univariate RG density are well-known in closed form (i.e. Eqs. \eqref{eq:mean diag} and \eqref{eq:approx diag}) and are used in the R-SBL DA algorithm.

\bibliographystyle{IEEEtran}
\bibliography{IEEEabrv,ManuscriptBib}

\begin{thebibliography}{10}
\providecommand{\url}[1]{#1}
\csname url@samestyle\endcsname
\providecommand{\newblock}{\relax}
\providecommand{\bibinfo}[2]{#2}
\providecommand{\BIBentrySTDinterwordspacing}{\spaceskip=0pt\relax}
\providecommand{\BIBentryALTinterwordstretchfactor}{4}
\providecommand{\BIBentryALTinterwordspacing}{\spaceskip=\fontdimen2\font plus
\BIBentryALTinterwordstretchfactor\fontdimen3\font minus
  \fontdimen4\font\relax}
\providecommand{\BIBforeignlanguage}[2]{{%
\expandafter\ifx\csname l@#1\endcsname\relax
\typeout{** WARNING: IEEEtran.bst: No hyphenation pattern has been}%
\typeout{** loaded for the language `#1'. Using the pattern for}%
\typeout{** the default language instead.}%
\else
\language=\csname l@#1\endcsname
\fi
#2}}
\providecommand{\BIBdecl}{\relax}
\BIBdecl

\bibitem{lawson1974solving}
C.~L. Lawson and R.~J. Hanson, \emph{Solving least squares problems}.\hskip 1em
  plus 0.5em minus 0.4em\relax SIAM, 1974, vol. 161.

\bibitem{jedynak2005maximum}
B.~M. Jedynak and S.~Khudanpur, ``{Maximum likelihood set for estimating a
  probability mass function},'' \emph{Neural Computation}, vol.~17, no.~7, pp.
  1508--1530, 2005.

\bibitem{peharz2012sparse}
R.~Peharz and F.~Pernkopf, ``Sparse nonnegative matrix factorization with
  $\ell$0-constraints,'' \emph{Neurocomputing}, vol.~80, pp. 38--46, 2012.

\bibitem{kim2007sparse}
H.~Kim and H.~Park, ``Sparse non-negative matrix factorizations via alternating
  non-negativity-constrained least squares for microarray data analysis,''
  \emph{Bioinformatics}, vol.~23, no.~12, pp. 1495--1502, 2007.

\bibitem{kim2008nonnegative}
------, ``Nonnegative matrix factorization based on alternating nonnegativity
  constrained least squares and active set method,'' \emph{SIAM Journal on
  Matrix Analysis and Applications}, vol.~30, no.~2, pp. 713--730, 2008.

\bibitem{fedorov2018unified}
I.~Fedorov, A.~Nalci, R.~Giri, B.~D. Rao, T.~Q. Nguyen, and H.~Garudadri, ``A
  unified framework for sparse non-negative least squares using multiplicative
  updates and the non-negative matrix factorization problem,'' \emph{Signal
  Processing}, 2018.

\bibitem{pauca2004text}
V.~P. Pauca, F.~Shahnaz, M.~W. Berry, and R.~J. Plemmons, ``Text mining rsing
  non-negative matrix factorizations,'' in \emph{Proceedings of the 2004 SIAM
  International Conference on Data Mining}, vol.~4, 2004, pp. 452--456.

\bibitem{monga2007robust}
V.~Monga and M.~K. Mih{\c{c}}ak, ``Robust and secure image hashing via
  non-negative matrix factorizations,'' \emph{IEEE Transactions on Information
  Forensics and Security}, vol.~2, no.~3, pp. 376--390, 2007.

\bibitem{loizou2005speech}
P.~C. Loizou, ``{Speech enhancement based on perceptually motivated Bayesian
  estimators of the magnitude spectrum},'' \emph{IEEE Transactions on Speech
  and Audio Processing}, vol.~13, no.~5, pp. 857--869, 2005.

\bibitem{fevotte2009nonnegative}
C.~F{\'e}votte, N.~Bertin, and J.-L. Durrieu, ``{Nonnegative matrix
  factorization with the Itakura-Saito divergence: With application to music
  analysis},'' \emph{Neural Computation}, vol.~21, no.~3, pp. 793--830, 2009.

\bibitem{sajda2004nonnegative}
P.~Sajda, S.~Du, T.~R. Brown, R.~Stoyanova, D.~C. Shungu, X.~Mao, and L.~C.
  Parra, ``Nonnegative matrix factorization for rapid recovery of constituent
  spectra in magnetic resonance chemical shift imaging of the brain,''
  \emph{IEEE Transactions on Medical Imaging}, vol.~23, no.~12, pp. 1453--1465,
  2004.

\bibitem{lin2006bayesian}
Y.~Lin and D.~D. Lee, ``Bayesian regularization and nonnegative deconvolution
  for room impulse response estimation,'' \emph{IEEE Transactions on Signal
  Processing}, vol.~54, no.~3, pp. 839--847, 2006.

\bibitem{potter2010sparsity}
L.~C. Potter, E.~Ertin, J.~T. Parker, and M.~Cetin, ``Sparsity and compressed
  sensing in radar imaging,'' \emph{Proceedings of the IEEE}, vol.~98, no.~6,
  pp. 1006--1020, 2010.

\bibitem{hurmalainen2012group}
A.~Hurmalainen, R.~Saeidi, and T.~Virtanen, ``{Group sparsity for speaker
  identity discrimination in factorisation-based speech recognition},'' in
  \emph{Interspeech}, 2012.

\bibitem{lustig2006kt}
M.~Lustig, J.~M. Santos, D.~L. Donoho, and J.~M. Pauly, ``{kt SPARSE: High
  frame rate dynamic MRI exploiting spatio-temporal sparsity},'' in
  \emph{Proceedings of the 13th Annual Meeting of ISMRM}, vol. 2420, 2006.

\bibitem{ghosh2013fiber}
A.~Ghosh, T.~Megherbi, F.~O. Boumghar, and R.~Deriche, ``{Fiber orientation
  distribution from non-negative sparse recovery},'' in \emph{10th
  International Symposium on Biomedical Imaging (ISBI)}.\hskip 1em plus 0.5em
  minus 0.4em\relax IEEE, 2013, pp. 254--257.

\bibitem{meng2011bayesian}
J.~Meng, J.~M. Zhang, Y.~Chen, and Y.~Huang, ``{Bayesian non-negative factor
  analysis for reconstructing transcription factor mediated regulatory
  networks},'' \emph{Proteome Science}, vol.~9, no.~1, p.~S9, 2011.

\bibitem{nalci:2016ISMRM}
A.~Nalci, B.~Rao, and T.~T. Liu, ``{Sparse Estimation of Quasi-periodic
  Spatiotemporal Components in Resting-State fMRI},'' in \emph{Proceedings of
  the 24th Annual Meeting of the ISMRM}, 2016, p. 3824.

\bibitem{liu2017global}
T.~T. Liu, A.~Nalci, and M.~Falahpour, ``The global signal in fmri: Nuisance or
  information?'' \emph{NeuroImage}, vol. 150, pp. 213--229, 2017.

\bibitem{aharon2005k}
M.~Aharon, M.~Elad, and A.~M. Bruckstein, ``{K-SVD and its non-negative variant
  for dictionary design},'' in \emph{Optics \& Photonics 2005}.\hskip 1em plus
  0.5em minus 0.4em\relax International Society for Optics and Photonics, 2005,
  pp. 591\,411--591\,411.

\bibitem{jiang2009note}
X.~Jiang and Y.~Ye, ``A note on complexity of lp minimization,''
  \emph{Preprint}, 2009.

\bibitem{eladbook}
M.~Elad, \emph{{Sparse and Redundant Representations}}.\hskip 1em plus 0.5em
  minus 0.4em\relax Springer New York, 2010.

\bibitem{tropp2007signal}
J.~A. Tropp and A.~C. Gilbert, ``Signal recovery from random measurements via
  orthogonal matching pursuit,'' \emph{IEEE Transactions on information
  theory}, vol.~53, no.~12, pp. 4655--4666, 2007.

\bibitem{needell2009cosamp}
D.~Needell and J.~A. Tropp, ``Cosamp: Iterative signal recovery from incomplete
  and inaccurate samples,'' \emph{Applied and Computational Harmonic Analysis},
  vol.~26, no.~3, pp. 301--321, 2009.

\bibitem{mallat1993matching}
S.~G. Mallat and Z.~Zhang, ``Matching pursuits with time-frequency
  dictionaries,'' \emph{IEEE Transactions on signal processing}, vol.~41,
  no.~12, pp. 3397--3415, 1993.

\bibitem{needell2009uniform}
D.~Needell and R.~Vershynin, ``Uniform uncertainty principle and signal
  recovery via regularized orthogonal matching pursuit,'' \emph{Foundations of
  computational mathematics}, vol.~9, no.~3, pp. 317--334, 2009.

\bibitem{c3}
A.~M. Bruckstein, D.~L. Donoho, and M.~Elad, ``{From sparse solutions of
  systems of equations to sparse modeling of signals and images},'' \emph{SIAM
  Review}, vol.~51, no.~1, pp. 34--81, 2009.

\bibitem{pati1993orthogonal}
Y.~C. Pati, R.~Rezaiifar, and P.~Krishnaprasad, ``{Orthogonal matching pursuit:
  Recursive function approximation with applications to wavelet
  decomposition},'' in \emph{Asilomar Conference on Signals, Systems and
  Computers}.\hskip 1em plus 0.5em minus 0.4em\relax IEEE, 1993, pp. 40--44.

\bibitem{efron2004least}
B.~Efron, T.~Hastie, I.~Johnstone, R.~Tibshirani \emph{et~al.}, ``Least angle
  regression,'' \emph{The Annals of statistics}, vol.~32, no.~2, pp. 407--499,
  2004.

\bibitem{figueiredo2007gradient}
M.~A. Figueiredo, R.~D. Nowak, and S.~J. Wright, ``Gradient projection for
  sparse reconstruction: Application to compressed sensing and other inverse
  problems,'' \emph{IEEE Journal of selected topics in signal processing},
  vol.~1, no.~4, pp. 586--597, 2007.

\bibitem{wright2009sparse}
S.~J. Wright, R.~D. Nowak, and M.~A. Figueiredo, ``Sparse reconstruction by
  separable approximation,'' \emph{IEEE Transactions on Signal Processing},
  vol.~57, no.~7, pp. 2479--2493, 2009.

\bibitem{donoho2005sparse}
D.~L. Donoho and J.~Tanner, ``Sparse nonnegative solution of underdetermined
  linear equations by linear programming,'' \emph{Proceedings of the National
  Academy of Sciences of the United States of America}, vol. 102, no.~27, pp.
  9446--9451, 2005.

\bibitem{nocedal2006numerical}
J.~Nocedal and S.~Wright, \emph{{Numerical Optimization}}.\hskip 1em plus 0.5em
  minus 0.4em\relax Springer Science \& Business Media, 2006.

\bibitem{boyd2004convex}
S.~Boyd and L.~Vandenberghe, \emph{{Convex Optimization}}.\hskip 1em plus 0.5em
  minus 0.4em\relax Cambridge University Press, 2004.

\bibitem{khajehnejad2011sparse}
M.~A. Khajehnejad, A.~G. Dimakis, W.~Xu, and B.~Hassibi, ``Sparse recovery of
  nonnegative signals with minimal expansion,'' \emph{IEEE Transactions on
  Signal Processing}, vol.~59, no.~1, pp. 196--208, 2011.

\bibitem{lin2007projected}
C.-b. Lin, ``Projected gradient methods for nonnegative matrix factorization,''
  \emph{Neural Computation}, vol.~19, no.~10, pp. 2756--2779, 2007.

\bibitem{grady2008compressive}
P.~D. Grady and S.~T. Rickard, ``{Compressive sampling of non-negative
  signals},'' in \emph{IEEE Workshop on Machine Learning for Signal
  Processing}.\hskip 1em plus 0.5em minus 0.4em\relax IEEE, 2008, pp. 133--138.

\bibitem{chartrand2008iteratively}
R.~Chartrand and W.~Yin, ``{Iteratively reweighted algorithms for compressive
  sensing},'' in \emph{IEEE International Conference on Acoustics, Speech and
  Signal Processing (ICASSP)}.\hskip 1em plus 0.5em minus 0.4em\relax IEEE,
  2008, pp. 3869--3872.

\bibitem{giri2015type}
R.~Giri and B.~D. Rao, ``{Type I and Type II Bayesian Methods for Sparse Signal
  Recovery using Scale Mixtures},'' \emph{IEEE Transactions on Signal
  Processing}, vol.~64, 2016.

\bibitem{tipping2001sparse}
M.~E. Tipping, ``{Sparse Bayesian learning and the relevance vector machine},''
  \emph{The Journal of Machine Learning Research}, vol.~1, pp. 211--244, 2001.

\bibitem{babacan2010bayesian}
S.~D. Babacan, R.~Molina, and A.~K. Katsaggelos, ``{Bayesian compressive
  sensing using Laplace priors},'' \emph{IEEE Transactions on Image
  Processing}, vol.~19, no.~1, pp. 53--63, 2010.

\bibitem{ji2008bayesian}
S.~Ji, Y.~Xue, and L.~Carin, ``{Bayesian compressive sensing},'' \emph{IEEE
  Transactions on Signal Processing}, vol.~56, no.~6, pp. 2346--2356, 2008.

\bibitem{andrews1974scale}
D.~F. Andrews and C.~L. Mallows, ``Scale mixtures of normal distributions,''
  \emph{Journal of the Royal Statistical Society. Series B (Methodological)},
  pp. 99--102, 1974.

\bibitem{wipf2007empirical}
D.~P. Wipf and B.~D. Rao, ``{An empirical Bayesian strategy for solving the
  simultaneous sparse approximation problem},'' \emph{IEEE Transactions on
  Signal Processing}, vol.~55, no.~7, pp. 3704--3716, 2007.

\bibitem{palmer2006variational}
J.~A. Palmer, ``{Variational and scale mixture representations of non-Gaussian
  densities for estimation in the Bayesian linear model: Sparse coding,
  independent component analysis, and minimum entropy segmentation},'' Ph.D.
  dissertation, University of California, San Diego, 2006.

\bibitem{palmer2005variational}
J.~Palmer, K.~Kreutz-Delgado, B.~D. Rao, and D.~P. Wipf, ``{Variational EM
  algorithms for non-Gaussian latent variable models},'' in \emph{Advances in
  Neural Information Processing Systems}, 2005, pp. 1059--1066.

\bibitem{lange1993normal}
K.~Lange and J.~S. Sinsheimer, ``Normal/independent distributions and their
  applications in robust regression,'' \emph{Journal of Computational and
  Graphical Statistics}, vol.~2, no.~2, pp. 175--198, 1993.

\bibitem{dempster1980iteratively}
A.~P. Dempster, N.~M. Laird, and D.~B. Rubin, ``{Iteratively reweighted least
  squares for linear regression when errors are normal/independent
  distributed},'' \emph{Multivariate Analysis V}, pp. 35--57, 1980.

\bibitem{dempster1977maximum}
------, ``{Maximum likelihood from incomplete data via the EM algorithm},''
  \emph{Journal of the Royal Statistical Society. Series B}, pp. 1--38, 1977.

\bibitem{wipf2011latent}
D.~P. Wipf, B.~D. Rao, and S.~Nagarajan, ``{Latent variable Bayesian models for
  promoting sparsity},'' \emph{IEEE Transactions on Information Theory},
  vol.~57, no.~9, pp. 6236--6255, 2011.

\bibitem{al2018gamp}
M.~Al-Shoukairi, P.~Schniter, and B.~D. Rao, ``{A GAMP-based low complexity
  sparse Bayesian learning algorithm},'' \emph{IEEE Transactions on Signal
  Processing}, vol.~66, no.~2, pp. 294--308, 2018.

\bibitem{c6}
D.~P. Wipf and B.~D. Rao, ``{Sparse Bayesian learning for basis selection},''
  \emph{IEEE Transactions on Signal Processing}, vol.~52, no.~8, pp.
  2153--2164, 2004.

\bibitem{zhang2011sparse}
L.~Zhang, M.~Yang, and X.~Feng, ``Sparse representation or collaborative
  representation: Which helps face recognition?'' in \emph{IEEE International
  Conference on Computer Vision (ICCV)}.\hskip 1em plus 0.5em minus 0.4em\relax
  IEEE, 2011, pp. 471--478.

\bibitem{fedorov2017multimodal}
I.~Fedorov, B.~D. Rao, and T.~Q. Nguyen, ``{Multimodal sparse Bayesian
  dictionary learning applied to multimodal data classification},'' in
  \emph{IEEE International Conference on Acoustics, Speech and Signal
  Processing (ICASSP)}.\hskip 1em plus 0.5em minus 0.4em\relax IEEE, 2017, pp.
  2237--2241.

\bibitem{c1}
M.~Harva and A.~Kab{\'a}n, ``Variational learning for rectified factor
  analysis,'' \emph{Signal Processing}, vol.~87, no.~3, pp. 509--527, 2007.

\bibitem{c2}
J.~W. Miskin, ``{Ensemble learning for independent component analysis},'' in
  \emph{Advances in Independent Component Analysis}.\hskip 1em plus 0.5em minus
  0.4em\relax Citeseer, 2000.

\bibitem{figueiredo2003adaptive}
M.~A. Figueiredo, ``{Adaptive sparseness for supervised learning},'' \emph{IEEE
  Transactions on Pattern Analysis and Machine Intelligence}, vol.~25, no.~9,
  pp. 1150--1159, 2003.

\bibitem{vila2014empirical}
J.~P. Vila and P.~Schniter, ``{An empirical-Bayes approach to recovering
  linearly constrained non-negative sparse signals},'' \emph{IEEE Transactions
  on Signal Processing}, vol.~62, no.~18, pp. 4689--4703, 2014.

\bibitem{gauvain1994maximum}
J.-L. Gauvain and C.-H. Lee, ``Maximum a posteriori estimation for multivariate
  gaussian mixture observations of markov chains,'' \emph{IEEE Transactions on
  Speech and Audio Processing}, vol.~2, no.~2, pp. 291--298, 1994.

\bibitem{bishop2006pattern}
C.~M. Bishop, ``{Pattern Recognition},'' \emph{Machine Learning}, 2006.

\bibitem{c8}
Z.~Zhang, T.-P. Jung, S.~Makeig, Z.~Pi, and B.~Rao, ``{Spatiotemporal sparse
  Bayesian learning with applications to compressed sensing of multichannel
  physiological signals},'' \emph{IEEE Transactions on Neural Systems and
  Rehabilitation Engineering}, vol.~22, no.~6, pp. 1186--1197, 2014.

\bibitem{horrace2005some}
W.~C. Horrace, ``Some results on the multivariate truncated normal
  distribution,'' \emph{Journal of Multivariate Analysis}, vol.~94, no.~1, pp.
  209--221, 2005.

\bibitem{robert2013monte}
C.~Robert and G.~Casella, \emph{{Monte Carlo statistical methods}}.\hskip 1em
  plus 0.5em minus 0.4em\relax Springer Science \& Business Media, 2013.

\bibitem{duane1987hybrid}
S.~Duane, A.~D. Kennedy, B.~J. Pendleton, and D.~Roweth, ``{Hybrid Monte
  Carlo},'' \emph{Physics Letters B}, vol. 195, no.~2, pp. 216--222, 1987.

\bibitem{pakman2014exact}
A.~Pakman and L.~Paninski, ``{Exact Hamiltonian Monte Carlo for truncated
  multivariate Gaussians},'' \emph{Journal of Computational and Graphical
  Statistics}, vol.~23, no.~2, pp. 518--542, 2014.

\bibitem{andrieu2003introduction}
C.~Andrieu, N.~De~Freitas, A.~Doucet, and M.~I. Jordan, ``{An introduction to
  MCMC for machine learning},'' \emph{Machine Learning}, vol.~50, no. 1-2, pp.
  5--43, 2003.

\bibitem{sherman1999conditions}
R.~P. Sherman, Y.-Y.~K. Ho, and S.~R. Dalal, ``{Conditions for convergence of
  Monte Carlo EM sequences with an application to product diffusion
  modeling},'' \emph{The Econometrics Journal}, vol.~2, no.~2, pp. 248--267,
  1999.

\bibitem{geman1984stochastic}
S.~Geman and D.~Geman, ``{Stochastic relaxation, Gibbs distributions, and the
  Bayesian restoration of images},'' \emph{IEEE Transactions on Pattern
  Analysis and Machine Intelligence}, no.~6, pp. 721--741, 1984.

\bibitem{li2015efficient}
Y.~Li and S.~K. Ghosh, ``{Efficient sampling methods for truncated multivariate
  normal and Student-t distributions subject to linear inequality
  constraints},'' \emph{Journal of Statistical Theory and Practice}, vol.~9,
  no.~4, pp. 712--732, 2015.

\bibitem{neath2013convergence}
R.~C. Neath \emph{et~al.}, ``{On convergence properties of the Monte Carlo EM
  algorithm},'' in \emph{Advances in Modern Statistical Theory and
  Applications}.\hskip 1em plus 0.5em minus 0.4em\relax Institute of
  Mathematical Statistics, 2013, pp. 43--62.

\bibitem{bickel2008covariance}
P.~J. Bickel and E.~Levina, ``{Covariance regularization by thresholding},''
  \emph{The Annals of Statistics}, pp. 2577--2604, 2008.

\bibitem{tong2014estimation}
T.~Tong, C.~Wang, and Y.~Wang, ``{Estimation of variances and covariances for
  high-dimensional data: a selective review},'' \emph{Wiley Interdisciplinary
  Reviews: Computational Statistics}, vol.~6, no.~4, pp. 255--264, 2014.

\bibitem{celeux1992stochastic}
G.~Celeux and J.~Diebolt, ``{A stochastic approximation type EM algorithm for
  the mixture problem},'' \emph{Stochastics: An International Journal of
  Probability and Stochastic Processes}, vol.~41, no. 1-2, pp. 119--134, 1992.

\bibitem{fisher2011improved}
T.~J. Fisher and X.~Sun, ``{Improved Stein-type shrinkage estimators for the
  high-dimensional multivariate normal covariance matrix},''
  \emph{Computational Statistics \& Data Analysis}, vol.~55, no.~5, pp.
  1909--1918, 2011.

\bibitem{ledoit2004well}
O.~Ledoit and M.~Wolf, ``{A well-conditioned estimator for large-dimensional
  covariance matrices},'' \emph{Journal of Multivariate Analysis}, vol.~88,
  no.~2, pp. 365--411, 2004.

\bibitem{kailath2000linear}
T.~Kailath, A.~H. Sayed, and B.~Hassibi, \emph{{Linear Estimation}}.\hskip 1em
  plus 0.5em minus 0.4em\relax Prentice Hall Upper Saddle River, NJ, 2000,
  vol.~1.

\bibitem{miskin2000ensemble}
J.~W. Miskin, ``{Ensemble learning for independent component analysis},'' in
  \emph{Advances in Independent Component Analysis}.\hskip 1em plus 0.5em minus
  0.4em\relax Citeseer, 2000.

\bibitem{rangan2011generalized}
S.~Rangan, ``{Generalized approximate message passing for estimation with
  random linear mixing},'' in \emph{International Symposium on Information
  Theory Proceedings (ISIT)}.\hskip 1em plus 0.5em minus 0.4em\relax IEEE,
  2011, pp. 2168--2172.

\bibitem{rangan2014convergence}
S.~Rangan, P.~Schniter, and A.~Fletcher, ``On the convergence of approximate
  message passing with arbitrary matrices,'' in \emph{IEEE International
  Symposium on Information Theory (ISIT)}.\hskip 1em plus 0.5em minus
  0.4em\relax IEEE, 2014, pp. 236--240.

\bibitem{caltagirone2014convergence}
F.~Caltagirone, L.~Zdeborov{\'a}, and F.~Krzakala, ``On convergence of
  approximate message passing,'' in \emph{IEEE International Symposium on
  Information Theory (ISIT)}.\hskip 1em plus 0.5em minus 0.4em\relax IEEE,
  2014, pp. 1812--1816.

\bibitem{Schniter_conv}
S.~Rangan, P.~Schniter, and A.~Fletcher, ``On the convergence of approximate
  message passing with arbitrary matrices,'' in \emph{IEEE International
  Symposium on Information Theory (ISIT)}, June 2014, pp. 236--240.

\bibitem{javanmard2013state}
A.~Javanmard and A.~Montanari, ``State evolution for general approximate
  message passing algorithms with applications to spatial coupling,''
  \emph{Information and Inference}, p. iat004, 2013.

\bibitem{vila2013expectation}
J.~P. Vila and P.~Schniter, ``{Expectation-maximization Gaussian-mixture
  approximate message passing},'' \emph{IEEE Transactions on Signal
  Processing}, vol.~61, no.~19, pp. 4658--4672, 2013.

\bibitem{horrace2005ranking}
W.~C. Horrace, ``On ranking and selection from independent truncated normal
  distributions,'' \emph{Journal of Econometrics}, vol. 126, no.~2, pp.
  335--354, 2005.

\bibitem{magdon2010approximating}
M.~Magdon-Ismail and J.~T. Purnell, ``{Approximating the covariance matrix of
  GMMs with low-rank perturbations},'' in \emph{Intelligent Data Engineering
  and Automated Learning}.\hskip 1em plus 0.5em minus 0.4em\relax Springer,
  2010, pp. 300--307.

\bibitem{liu2009slep}
J.~Liu, S.~Ji, J.~Ye \emph{et~al.}, ``{SLEP: Sparse learning with efficient
  projections},'' \emph{Arizona State University}, vol.~6, p. 491, 2009.

\bibitem{bruckstein2008uniqueness}
A.~M. Bruckstein, M.~Elad, and M.~Zibulevsky, ``On the uniqueness of
  nonnegative sparse solutions to underdetermined systems of equations,''
  \emph{IEEE Transactions on Information Theory}, vol.~54, no.~11, pp.
  4813--4820, 2008.

\bibitem{candes2006stable}
E.~J. Candes, J.~K. Romberg, and T.~Tao, ``Stable signal recovery from
  incomplete and inaccurate measurements,'' \emph{Communications on Pure and
  Applied Mathematics}, vol.~59, no.~8, pp. 1207--1223, 2006.

\bibitem{vila2015adaptive}
J.~Vila, P.~Schniter, S.~Rangan, F.~Krzakala, and L.~Zdeborov{\'a}, ``Adaptive
  damping and mean removal for the generalized approximate message passing
  algorithm,'' in \emph{Acoustics, Speech and Signal Processing (ICASSP), 2015
  IEEE International Conference on}.\hskip 1em plus 0.5em minus 0.4em\relax
  IEEE, 2015, pp. 2021--2025.

\bibitem{li2013robust}
T.~Li and Z.~Zhang, ``Robust face recognition via block sparse bayesian
  learning,'' \emph{Mathematical Problems in Engineering}, vol. 2013, 2013.

\bibitem{he2013two}
R.~He, W.-S. Zheng, B.-G. Hu, and X.-W. Kong, ``Two-stage nonnegative sparse
  representation for large-scale face recognition,'' \emph{IEEE Transactions on
  Neural Networks and Learning Systems}, vol.~24, no.~1, pp. 35--46, 2013.

\bibitem{he2011maximum}
R.~He, W.-S. Zheng, and B.-G. Hu, ``Maximum correntropy criterion for robust
  face recognition,'' \emph{IEEE Transactions on Pattern Analysis and Machine
  Intelligence}, vol.~33, no.~8, pp. 1561--1576, 2011.

\bibitem{vo2009nonnegative}
N.~Vo, B.~Moran, and S.~Challa, ``Nonnegative-least-square classifier for face
  recognition,'' \emph{Advances in Neural Networks--ISNN 2009}, pp. 449--456,
  2009.

\bibitem{wright2009robust}
J.~Wright, A.~Y. Yang, A.~Ganesh, S.~S. Sastry, and Y.~Ma, ``Robust face
  recognition via sparse representation,'' \emph{IEEE Transactions on Pattern
  Analysis and Machine Intelligence}, vol.~31, no.~2, pp. 210--227, 2009.

\bibitem{turk1991eigenfaces}
M.~Turk and A.~Pentland, ``Eigenfaces for recognition,'' \emph{Journal of
  Cognitive Neuroscience}, vol.~3, no.~1, pp. 71--86, 1991.

\bibitem{he2005face}
X.~He, S.~Yan, Y.~Hu, P.~Niyogi, and H.-J. Zhang, ``Face recognition using
  laplacianfaces,'' \emph{IEEE Transactions on Pattern Analysis and Machine
  Intelligence}, vol.~27, no.~3, pp. 328--340, 2005.

\bibitem{martinez1998ar}
A.~M. Martinez, ``{The AR face database},'' \emph{CVC technical report}, 1998.

\end{thebibliography}

\begin{figure}[h!] 
\vspace{-2.6em}
\begin{IEEEbiography}[{\includegraphics[width=1in,height=1.25in,clip,keepaspectratio]{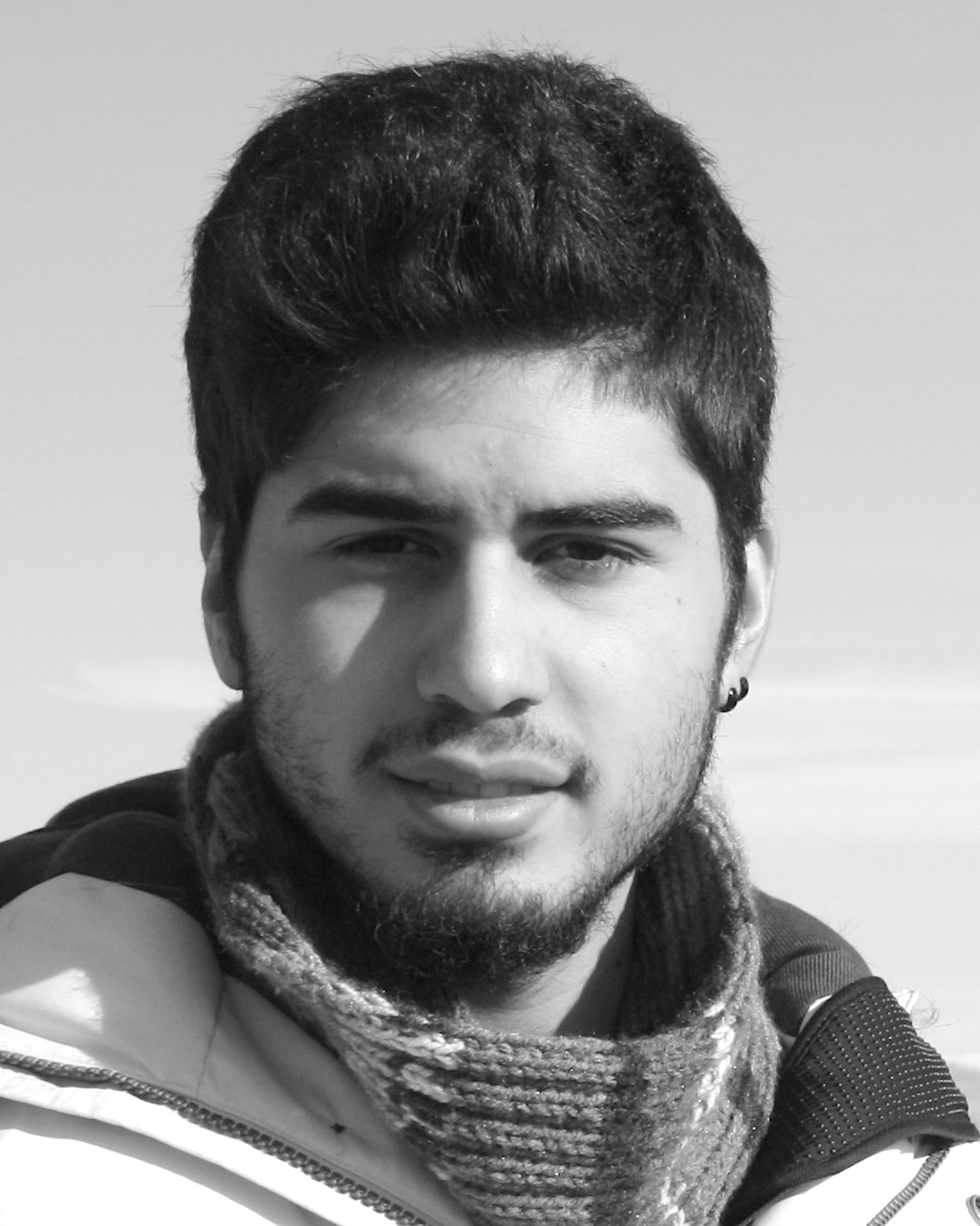}}]{Alican Nalci} (S'15) received the B.Sc. degree in Electrical and Electronics Engineering from Bilkent University, Ankara, Turkey, in 2013. 
He received the M.Sc. degree in Electrical and Computer Engineering from the University of California San Diego, La Jolla, CA, USA, in 2015, where he is currently pursuing a Ph.D. degree. 
His research interests include sparse signal recovery, machine learning, signal processing and functional magnetic resonance imaging (fMRI).
\end{IEEEbiography} 
\vspace{-2.6em}
\begin{IEEEbiography}[{\includegraphics[width=1in,height=1.25in,clip,keepaspectratio]{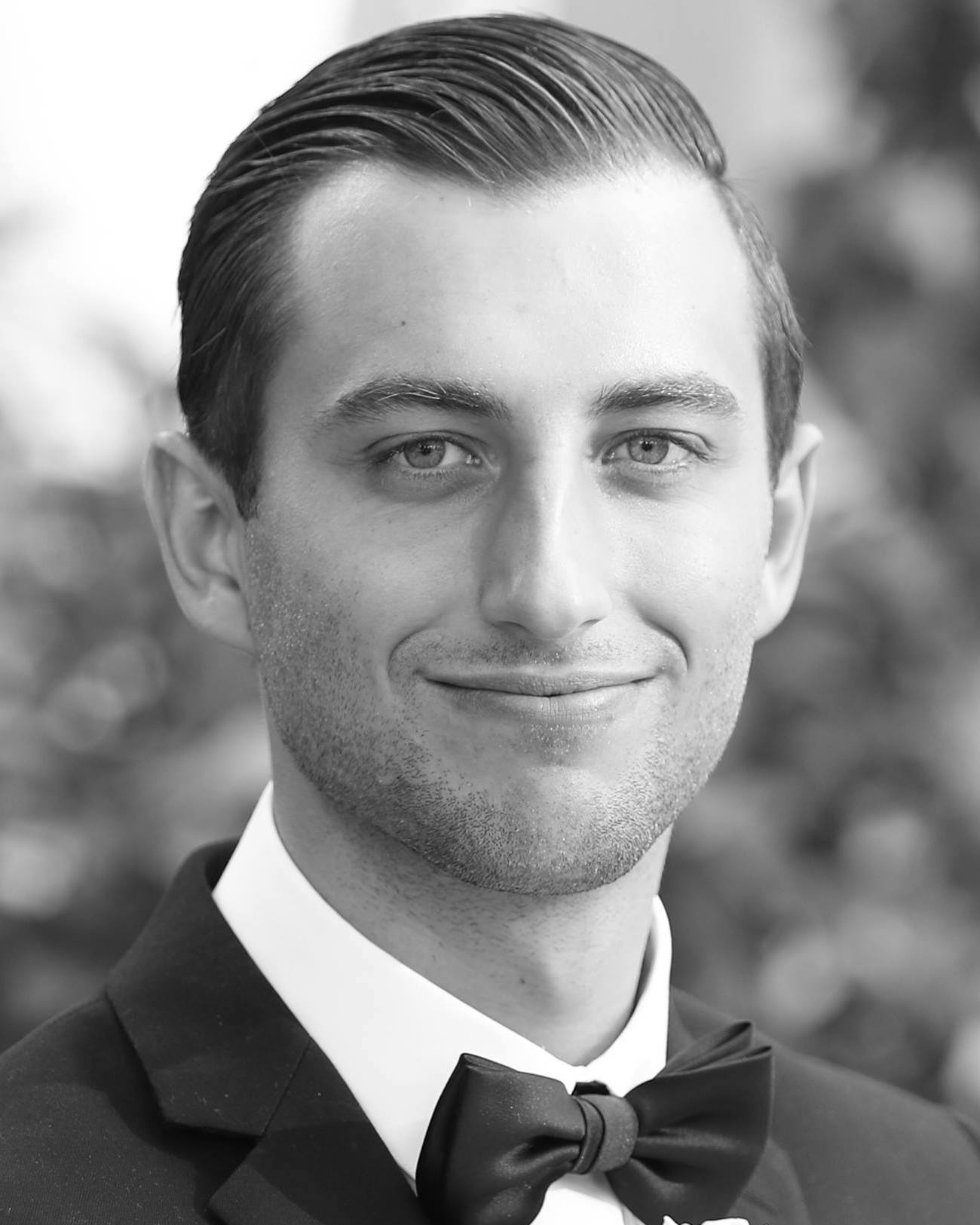}}]{Igor Fedorov}
(S'15) receieved the B.Sc. and M.Sc. degrees in Electrical Engineering from the University of Illinois at Urbana-Champaign in 2012 and 2014, respectively. 
He is currently pursuing a Ph.D. degree in Electrical Engineering at the University of California San Diego, La Jolla, CA, USA. 
His research interests include sparse signal recovery, machine learning, and signal processing.
\end{IEEEbiography} 
\vspace{-2.6em}
\begin{IEEEbiography}[{\includegraphics[width=1in,height=1.25in,clip,keepaspectratio]{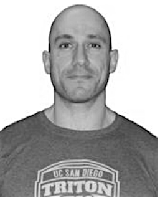}}]{Maher Al-Shoukairi}
(S'17) received his B.Sc. degree in Electrical Engineering from the University of Jordan in 2005. He received his M.Sc. degree in Electrical Engineering from Texas A\&M University in 2008, after which he joined Qualcomm Inc. on the same year to date. He is currently pursuing his Ph.D. degree in Electrical Engineering at the University of California, San Diego.
\end{IEEEbiography} 
\vspace{-2.6em}
\begin{IEEEbiography}[{\includegraphics[width=1in,height=1.25in,clip,keepaspectratio]{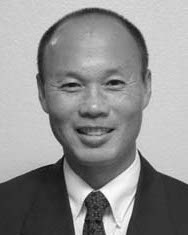}}]{Thomas T. Liu} received the B.S degree in Electrical Engineering from the Massachusetts Institute of Technology, Cambridge, MA, USA,  and the M.S. and Ph.D. degrees from Stanford University,  Stanford CA, USA,  in 1988, 1993, and 1999, respectively.  Since 1999 he has been with the University of California San Diego, La Jolla, CA, USA,  where he is currently a Professor in the Departments of Radiology, Psychiatry, and Bioengineering and Director of the UCSD Center for Functional MRI. 
\end{IEEEbiography} 
\vspace{-2.6em}
\begin{IEEEbiography}[{\includegraphics[width=1in,height=1.25in,clip,keepaspectratio]{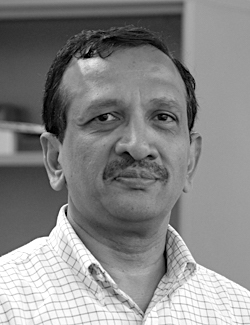}}]{Bhaskar D. Rao} (S'80-M'83-SM'91-F'00)
received the B.Tech. degree in electronics and electrical communication engineering from the Indian Institute of Technology of Kharagpur, Kharagpur, India, in 1979, and the M.S. and Ph.D. degrees from the University of Southern California, Los Angeles, CA, USA, in 1981 and 1983, respectively. Since 1983, he has been at the University of California at San Diego, San Diego, CA, USA, where he is currently a Distinguished Professor in the Department of Electrical and Computer Engineering. 
\end{IEEEbiography}
\vspace{-2.6em}
\end{figure}

\end{document}